\documentclass[sigconf]{acmart}

\copyrightyear{2025}
\acmYear{2025}
\setcopyright{cc}
\setcctype{by-nc-sa}
\acmConference[SIGIR '25]{Proceedings of the 48th International ACM SIGIR Conference on Research and Development in Information Retrieval}{July 13--18, 2025}{Padua, Italy}
\acmBooktitle{Proceedings of the 48th International ACM SIGIR Conference on Research and Development in Information Retrieval (SIGIR '25), July 13--18, 2025, Padua, Italy}\acmDOI{10.1145/3726302.3729959}
\acmISBN{979-8-4007-1592-1/2025/07}

\AtBeginDocument{%
  \providecommand\BibTeX{{%
    \normalfont B\kern-0.5em{\scshape i\kern-0.25em b}\kern-0.8em\TeX}}}

\usepackage{balance}
\usepackage{multirow}
\usepackage{array}
\usepackage{tabularx,booktabs}
\newcolumntype{Y}{>{\centering\arraybackslash}X}
\usepackage{enumitem}
\newcommand{\proposed}{\textsf{DITTO}}
\usepackage{xcolor}
\usepackage{pifont}
\usepackage{makecell}

\begin{document}

\title{Dynamic Time-aware Continual User Representation Learning}

\author{Seungyoon Choi}
\affiliation{%
  \institution{KAIST}
  \city{Daejeon}
  \country{Republic of Korea}}
\email{csyoon08@kaist.ac.kr}

\author{Sein Kim}
\affiliation{%
  \institution{KAIST}
  \city{Daejeon}
  \country{Republic of Korea}}
\email{rlatpdlsgns@kaist.ac.kr}

\author{Hongseok Kang}
\affiliation{%
  \institution{KAIST}
  \city{Daejeon}
  \country{Republic of Korea}}
\email{ghdtjr0311@kaist.ac.kr}

\author{Wonjoong Kim}
\affiliation{%
  \institution{KAIST}
  \city{Daejeon}
  \country{Republic of Korea}}
\email{wjkim@kaist.ac.kr}

\author{Chanyoung Park}
\authornote{Corresponding author.}
\affiliation{%
  \institution{KAIST}
  \city{Daejeon}
  \country{Republic of Korea}}
\email{cy.park@kaist.ac.kr}

\renewcommand{\shortauthors}{Seungyoon Choi, Sein Kim, Hongseok Kang, Wonjoong Kim, \& Chanyoung Park}

\begin{abstract}

Traditional user modeling (UM) approaches have primarily focused on designing models for a single specific task, but they face limitations in generalization and adaptability across various tasks. Recognizing these challenges, recent studies have shifted towards continual learning (CL)-based universal user representation learning aiming to develop a single model capable of handling multiple tasks. 
Despite advancements, existing methods are in fact evaluated under an unrealistic scenario that does not consider the passage of time as tasks progress, which overlooks newly emerged items that may change the item distribution of previous tasks.
In this paper, we introduce a practical evaluation scenario on which CL-based universal user representation learning approaches should be evaluated, which takes into account the passage of time as
tasks progress. 
Then, we propose a novel framework \textsf{\textbf{D}}ynam\textsf{\textbf{I}}c \textsf{\textbf{T}}ime-aware con\textsf{\textbf{T}}inual user representati\textsf{\textbf{O}}n learner, named \proposed, designed to alleviate catastrophic forgetting despite continuous shifts in item distribution, while also allowing the knowledge acquired from previous tasks to adapt to the current shifted item distribution.
Through our extensive experiments, we demonstrate the superiority of~\proposed~over state-of-the-art methods under a practical evaluation scenario. Our source code is available at \url{https://github.com/seungyoon-Choi/DITTO_official}.
\end{abstract}



\maketitle

\section{Introduction}
Recent research emphasizes the importance of user modeling (UM) in delivering personalized services to individual customers \cite{10.1145/2806416.2806527,5197422,hidasi2015session,kang2018self,yuan2019simple, kim2023melt}. By learning reliable user representations, UM enables the provision of tailored recommendations and accurate predictions of user behavior. Understanding customer interests and preferences not only enhances service quality but also fosters the creation of products and services that better resonate with users.

Most existing studies on UM focus on designing models for a single specific task \cite{guo2017deepfm,yuan2019simple,zhou2019deep}, e.g., a model for click prediction and another model for user profiling. However, since enterprises offer multiple services to users, maintaining an independent model for each service entails the cost for maintaining and updating the model, and this becomes especially impractical as new services are continuously launched.
To this end, research on universal user representation learning has recently garnered attention \cite{yuan2020parameter, yuan2021one, Ni2018PerceiveYU, gu2021exploiting}. Universal user representation refers to a user representation that can be generally applied to multiple tasks, representing a user in a universal manner. The core of universal user representation learning lies in training a single user model applicable across various tasks.

\looseness=-1
Prior studies have employed multi-task learning (MTL)  \cite{crawshaw2020multi,ruder2017overview} to learn universal user representation, whose key idea is to simultaneously train multiple tasks aiming to maximize/minimize positive/negative transfer between tasks by understanding the relationship between tasks, thereby enhancing the performance of all tasks \cite{ni2018perceive,10.1145/3219819.3220007}. However, MTL is impractical as it requires all user data to be simultaneously available for all tasks at once.
To address these issues,  transfer learning (TL) has been introduced, utilizing knowledge learned from models pre-trained on source tasks to tackle a target task \cite{yuan2020parameter,10.5555/2891460.2891628,yang2021autoft}. However, TL always transfers knowledge between a pair of tasks, i.e., from a source task to a target task, and thus is not suitable for the essence of universal user representation learning, which aims to learn a universal model that can be adapted to a more diverse range of tasks. 
For this reason, recent studies mainly focus on continual learning (CL), whose key idea is to \textit{sequentially train various tasks} with a single model, while preventing 
catastrophic forgetting of the knowledge acquired from earlier tasks \cite{kumari2022retrospective,zhou2021overcoming,yoon2017lifelong}. 

\begin{table}[t]
  \caption{The number of new items emerged over a time horizon.}
  \label{tab: new_item_stat}
  \centering
  \resizebox{1.0\columnwidth}{!}{
  \begin{tabular}{c||c|cccccc}
    \toprule
     \multirow{2}{*}{\makecell{Tmall\\Dataset}} & Num. items & \multicolumn{6}{c}{The number of new items emerged} \\ \cline{3-8} 
     
     & \multicolumn{1}{c|}{given on 8/11} & 8/11 \textasciitilde~8/26 & 8/26 \textasciitilde~9/11 & 9/11 \textasciitilde~9/26 & 9/26 \textasciitilde~10/11 & 10/11 \textasciitilde~10/26 &10/26\textasciitilde~11/12 \\\midrule\midrule
Click & 570.6K & 65.0K & 79.0K & 61.0K & 58.9K & 77.0K & 171.3K  \\ 
Cart    & 6.2K & 1.1K & 1.9K & 1.7K & 1.9K & 3.2K & 27.1K  \\
Purchase     & 153.3K & 16.8K & 26.5K & 19.2K & 18.4K & 21.3K & 117.3K  \\ 
Favorite     & 195.2K & 27.2K & 34.4K & 28.1K & 28.2K & 39.1K & 93.1K  \\ \bottomrule
\end{tabular}
}
\end{table}

Although existing CL-based universal user representation learning approaches have achieved promising results, we argue that their evaluation scenario is unrealistic: \textit{it overlooks the passage of time as tasks progress}. In other words, while Task 2 is being trained, Task 1 is assumed to remain static, ignoring the potential introduction of new items to Task 1 (please refer Fig.~\ref{fig:setting}). 
However, as shown in Table \ref{tab: new_item_stat},  new items continuously emerge over a time horizon. For example, for the click prediction task, the number of items given on 8/11 is 570.6K, while 65K new items emerged within the next two weeks (i.e., 8/11 \textasciitilde~8/26). 
This implies that tasks are likely to encounter a continuous shift in the item distribution as time passes, and we argue that this should not be overlooked when evaluating the CL-based universal user representation approaches.

\begin{figure}[t]
  \centering
  \includegraphics[width=0.7\linewidth]{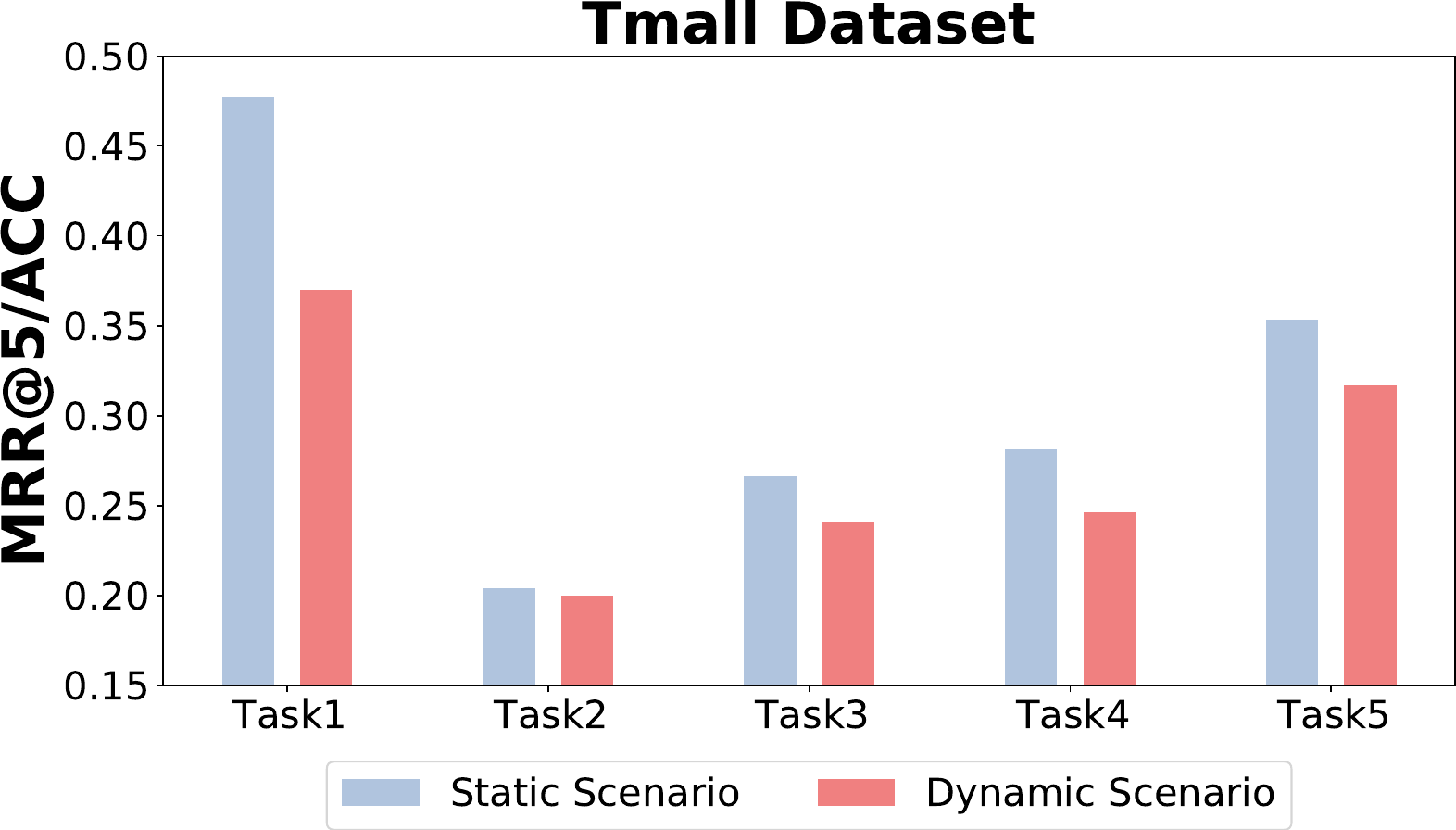}
  \caption[Caption for LOF]
  {Performance of TERACON in scenarios where new items do not emerge (blue) and new items emerge (red) as the task progresses. The next item prediction tasks (Task1-4) are evaluated using MRR@5, while user profile prediction tasks (Task5) is evaluated using Accuracy\footnotemark[1].}
  \label{fig:teracon}
\end{figure}

On the other hand, as the existing state-of-the-art approach, TERACON~\cite{kim2023task}, is originally designed considering the unrealistic evaluation scenario described above, we observe that the performance of TERACON when new items emerge as tasks progress (i.e., Dynamic scenario) is significantly inferior to that when new items do not emerge (i.e., Static scenario) in Figure~\ref{fig:teracon}. We attribute this to two reasons: \textbf{(1) The retention module of TERACON induces negative transfer.} TERACON retains the knowledge learned from the previous task through a knowledge retention module that transfers the knowledge of the current model regarding the previous task to help train the current model itself. This is done by generating pseudo-labels of the previous task for the given user behavior sequences in the current task. However, for the user behavior sequence in the current task that interacts with items not seen during the previous task training, reliable pseudo-labels cannot be generated, and retaining such unreliable knowledge results in negative transfer. \textbf{(2) The previous task cannot adapt to the changed item distribution in the current task.} While successful knowledge retention in (1) is important, it is equally critical to adapt to shifted distribution. In dynamic scenarios, simply retaining knowledge from the previous task is insufficient; the retained knowledge must also be capable of adapting to the shifted distribution in the current task. Since TERACON (as well as all the existing CL-based universal user representation learning approaches \cite{yuan2021one,kim2023task}) is designed for scenarios where there is no time difference between training and inference, there is no mechanism to address distribution shifts.

\addtocounter{footnote}{1}
\footnotetext{Results on Task 6 is omitted as forgetting does not occur in the last task.}

\looseness=-1
In this work, we first introduce a practical evaluation scenario on which CL-based universal user representation learning approaches should be evaluated. Then, we propose a novel framework \textsf{\textbf{D}}ynam\textsf{\textbf{I}}c \textsf{\textbf{T}}ime-aware con\textsf{\textbf{T}}inual user representati\textsf{\textbf{O}}n learner, named \proposed, designed to alleviate catastrophic forgetting despite continuous shifts in item distribution, while also allowing the knowledge acquired from previous tasks to adapt to the current shifted item distribution.
The main idea is categorized into two aspects. 
First, we propose a distribution-aware forward knowledge transfer (FKT) module that alleviates catastrophic forgetting by generating reliable pseudo-labels from a previous task.
Specifically, for reliable pseudo-label generation, we rely only on the users of the current task whose behavior sequence distribution is similar to the distribution of item embeddings learned during  previous tasks.
Second, we propose a distribution-aware backward knowledge transfer (BKT) module that allows the previously acquired knowledge to adapt to the current shifted item distribution.
Specifically, this module samples users in the current task whose behavior sequence distribution has been shifted the most compared with that in the previous task, and use them to fine-tune the previously acquired knowledge so that it adapts to the current shifted item distribution. 
It is worth noting that while forward/backward transfer modules are not new in CL-based universal user representation learning, to the best of our knowledge, this is the first work that investigates how/what/why to forward/backward transfer under continuous shifts in item distribution.

Our contributions are summarized as follows:
\begin{itemize}[leftmargin = 3mm]

    \item In this work, we introduce a practical scenario on which CL-based universal user representation learning approaches should be evaluated, which takes into account the passage of time as tasks progress. To the best of our knowledge, this is the first study to revisit the evaluation scenario of CL-based universal user representation learning research.

    \item To tackle the limitation of existing approaches under the proposed evaluation scenario, we propose \proposed, which maximizes knowledge transfer in both the forward and backward directions, where the forward transfer alleviates catastrophic forgetting while the backward transfer allows the previously acquired knowledge to adapt to the current shifted item distribution.

    \item Extensive experiments under the practical evaluation scenario show that \proposed~outperforms state-of-the-art  CL-based universal user representation learning methods.

\end{itemize}

\section{Related Works}

\subsection{Universal User Representation}

The goal of User modeling (UM) is to the user profile, a conceptual understanding tailored for personalized recommender systems. At its core, UM aims to create a representation for each user, utilizing either the items that the user interacted with or the associated features. These representations serve various purposes, including tasks like response prediction and recommendation \cite{10.5555/3491440.3492135}. 
With recent advances in deep neural networks, researchers have shown a growing interest in neural network-based user modeling. This encompasses factorization-based \cite{guo2017deepfm, 10.1145/3077136.3080777}, recurrent neural network-based \cite{hidasi2015session}, and graph-based approaches \cite{wang2019neural,wu2019session}. Nevertheless, these UM methods often prioritize task-specific user representations, potentially neglecting the broader interests of users.

Recent studies mainly focus on continual learning (CL), whose key idea is to sequentially train various tasks with a single model, while preventing catastrophic forgetting of the knowledge acquired from earlier tasks. 
Most recently, TERACON \cite{kim2023task} captures task relationships through task embeddings, and achieves state-of-the-art performance in continual user representation learning by preventing catastrophic forgetting through knowledge retention.

While existing studies on CL-based universal user representation learning approaches have demonstrated superior performance, they are not suitable for real-world applications as they do not consider the passage of time as tasks progress. In this work, we introduce a practical scenario on which CL-based universal user representation learning approaches should be evaluated, which takes into account the passage of time as tasks progress.

\subsection{Continual Learning}

Continual learning is an approach in which a model gathers insights from an ongoing flow of datasets while preserving information acquired from previous tasks. Nevertheless, as the model advances through tasks, performance is frequently decreased due to the loss of knowledge gained from previous tasks, a phenomenon known as catastrophic forgetting \cite{mccloskey1989catastrophic,ratcliff1990connectionist}, and preventing it is the fundamental aim of continual learning.
The regularization-based approach, one of the three continual learning approaches, aims to regulate the model's parameters to minimize catastrophic forgetting while training on new tasks. This strategy focuses on preserving crucial weights that played a vital role in learning previous tasks, while permitting the adaptation and acquisition of new knowledge by the remaining weights \cite{kirkpatrick2017overcoming,mallya2018piggyback,mallya2018packnet}. 
HAT \cite{serra2018overcoming} trains a soft mask for the output of each layer and the parameters of the entire model are trained throughout the entire task sequence.
Inspired by HAT, TERACON~\cite{kim2023task} trains universal user representation by capturing relations between tasks by learning task-specific masks and retaining knowledge from previous tasks through pseudo-labeling. However, TERACON used a limited setting with static data without considering the progression of time as tasks proceed. 
Therefore, TERACON's knowledge retention module induces negative transfer due to distribution shifts and, furthermore, fails to effectively adapt the previously acquired knowledge to the current distribution.

\section{Preliminaries}

\subsection{Problem Formulation}
\label{sec:problem_formulation}

\subsubsection{\textbf{Notations}}
Let $\mathcal{T} = \left\{T_1, T_2, ..., T_i, ..., T_{M}\right\}$ represent the collection of consecutive tasks, which contains a set of next item prediction tasks, i.e., $\mathcal{T}_{item}\subset \mathcal{T}$, and a set of user profile prediction tasks, i.e., $\mathcal{T}_{profile}\subset \mathcal{T}$. Note that $\mathcal{T}_{item} \cup \mathcal{T}_{profile} = \mathcal{T}$.
We use $t=\{t_1, t_2, ..., t_j, ..., t_M\}$ to denote the timestamp, and in accordance with this, $T^{t_j}_i$ indicates task $T_i$ is trained or inferred at time $t_j$ (e.g., $T^{t_2}_4$ represents $T_4$ is trained or inferred at $t_2$)\footnote{
For simplicity, we omit the timestamp (i.e., we use $T_i$ instead of $T^{t_j}_i$) when referring to general context that does not require specifying a particular time point $t_j$.}.
$\mathcal{U} = \left\{u_1, u_2, ..., u_{N}\right\}$ denotes the set of users across entire tasks, with each user $u_l$ being characterized by their behavior sequence up to the time point $t_j$ $\mathbf{x}^{u_l}_{t_j}=\left\{x_1^{u_l},x_2^{u_l},...,x_{n}^{u_l}\right\}$, where $x^{u_l}_k \in \mathcal{I}$ denotes the $k$-th interaction in the interaction sequence of $u_l$, and $\mathcal{I}$ is the set of items. Due to variations in the length of behavior sequences for each user at different time points, we fix the length as $n$ and pad the sequence to the left until the length reaches $n$. As time passes (e.g., $t_j \rightarrow t_{j+1}$), the number of interacted items increases, resulting in an expansion of items within the set of $n$ items, and thus the number of padding decreases. For each task $T^{t_j}_i$, there is only a subset of users $\mathcal{U}^{T^{t_j}_i}=\left\{u_1, u_2, ..., u_{|\mathcal{U}^{T^{t_j}_i}|} \right\}$, among the entire set of users $\mathcal{U}$, i.e., $\mathcal{U}^{T^{t_j}_i} \subset \mathcal{U}$. Note that $\mathcal{U}^{T^{t_j}_i}$ is associated with the set of labels $\mathbf{Y}^{T^{t_j}_i} = \{y^{T^{t_j}_i}_{u_1}, y^{T^{t_j}_i}_{u_2}, ..., y^{T^{t_j}_i}_{u_{|\mathcal{U}^{T^{t_j}_i}|}} \}$, where $y^{T^{t_j}_i}_{u_l}$ denotes the label of user $u_l$ in task $T^{t_j}_i$ (e.g., favorite item id, purchased item id, and gender of user $u_l$). We use $\mathcal{Y}^{T^{t_j}_i}$ to represent the set of unique labels in $\mathbf{Y}^{T^{t_j}_i}$, and $\mathbf{y}_{u_l}^{T^{t_j}_i}\in\{0,1\}^{|\mathcal{Y}^{T^{t_j}_i}|}$ represents the one-hot transformation of $y^{T^{t_j}_i}_{u_l}$. For the case of the next item prediction task $T_i$, $\mathcal{Y}^{T^{t_j}_i}$, which is the set of next item candidates, is expanded over time due to the emergence of new items (i.e., $|\mathcal{Y}^{T^{t_j}_i}| < |\mathcal{Y}^{T^{t_{j+1}}_i}|$).

\subsubsection{\textbf{Task Description}}
There is a set of sequential tasks with different timestamps $\mathcal{T}^t = \left\{T^{t_1}_1, T^{t_2}_2, ..., T^{t_k}_k, ..., T^{t_M}_{M}\right\}$, that is, the first task $T_{1}$ started at $t_1$, the second task $T_{2}$ started at $t_2$, and so on. For each task $T^{t_j}_i$, we are given a user set $\mathcal{U}^{T^{t_j}_i}$ and a label set $\mathbf{Y}^{T^{t_j}_i}$. The objective is to sequentially train each task using a single model $\mathcal{M}$ and predict labels for each user $u_l \in \mathcal{U}^{T^{t_j}_i}$, i.e., $\mathbf{\hat{y}}_{u_l}^{T^{t_j}_i} = G^{T_i}(\mathcal{M}(\mathbf{x}^{u_l}_{t_j}))$, where $\mathcal{M}$ is the backbone network that generates universal user representations, and $G^{T_i}$ is a task-specific projector for task $T_i$.\footnote{Please note that, in the next item prediction task $T_i\in\mathcal{T}_{item}$, $G^{T_i}$ serves the role of projecting the universal representation of the user, and in the user profile prediction task $T_k\in\mathcal{T}_{profile}$, $G^{T_k}$ serves as a classifier.} After sequentially training on all tasks, we conduct inference on all tasks using a single network $\mathcal{M}$.


\subsection{Backbone Structure}
\label{sec:backbone_network}

\subsubsection{Backbone Network $\mathcal{M}$}
Although our proposed~\proposed~is a backbone network-agnostic model, to be consistent with prior studies \cite{yuan2021one,kim2023task}, we employ temporal convolutional network (TCN) \cite{yuan2019simple} as the backbone network. TCN learns the representation of user $u_l$ based on the sequence of user's interacted items, i.e., $\textbf{x}^{u_l}=\{x^{u_l}_1, x^{u_l}_2, ..., x^{u_l}_n\}$. Specifically, given an initial embedding matrix $\mathbf{E}^{u_l}_0 \in \mathbb{R}^{n\times f}$ corresponding to $\textbf{x}^{u_l}$, where $\mathbf{E}_0\in \mathbb{R}^{|\mathcal{I}|\times f}$ is initialized with item embedding matrix $\mathbf{I}\in\mathbb{R}^{|\mathcal{I}|\times f}$, and $f$ is the embedding dimension, we pass it to the TCN encoder, which is structured with a stack of residual blocks. Each block consists of two temporal convolutional layers and normalization layers. The $k$-th residual block is defined as:
\begin{equation}
    \mathbf{E}^{u_l}_k = F_k(\mathbf{E}^{u_l}_{k-1}) + \mathbf{E}^{u_l}_{k-1}, \,\,\,\,\text{where } l=1,...,K.
    \label{Eq 1}
\end{equation}
Here, $\mathbf{E}^{u_l}_{k-1}$ and $\mathbf{E}^{u_l}_k$ represent the input and output of residual block, and $F_k$ is the residual mapping to be learned as described below:
\begin{equation}
    F_k(\mathbf{E}^{u_l}_{k-1})=\sigma(\phi_2({LN}_2(\sigma(\phi_1({LN}_1(\mathbf{E}^{u_l}_{k-1})))))),
    \label{Eq 2}
\end{equation}
where $\sigma$ represents the ReLU function \cite{nair2010rectified}, $LN$ denotes layer normalization \cite{ba2016layer}, $\phi$ represents the TCN layer, and biases are omitted.
Note that given the output of the last layer $K$, i.e., $\mathbf{E}^{u_l}_K$, we consider its last row, i.e., $\mathbf{E}^{u_l}_K[-1,:]\in\mathbb{R}^f$, as the representation of user $u_l$.

\subsubsection{Task-specific Mask $\mathbf{m}$}
Similar to the prior works~\cite{serra2018overcoming,kim2023task} in continual learning, we adapt a trainable soft mask $\mathbf{m}$ to the backbone network as a task-specific mask to preserve previous task information while avoiding interference with current task learning. $\mathbf{m}_{k}^{T_i}\in\mathbb{R}^f$ denote the task-specific mask of task $T_i$ in layer $k$. Then, given the behavior sequence of user $u_l$, we apply the task-specific mask $\mathbf{m}_{k}^{T_i}$ to derive the task-specific output of $u_l$ in task $T_i$ as follows:
\begin{equation}
    \mathbf{E}^{u_l}_k = F_k(\mathbf{E}^{u_l}_{k-1};\mathbf{m}_{k}^{T_i}) + \mathbf{E}^{u_l}_{k-1},
\end{equation}
where $F_k(\mathbf{E}^{u_l}_{k-1};\mathbf{m}_{k}^{T_i})$ represents the masked residual mapping in layer $k$, derived by element-wise multiplying the mask $\mathbf{m}_{k}^{T_i}$ with each output of the ReLU activation in $F_k(\cdot)$. Let $\mathcal{M}$ represents the entire backbone network from layer 1 to $K$, the final output embedding of user $u_l$ in task $T_i$, i.e., $\mathbf{E}^{u_l}_K$ is defined as follows:

\begin{equation}
    \mathbf{E}^{u_l}_K = \mathcal{M}(\mathbf{x}^{u_l};\mathbf{m}^{T_i}).
\end{equation}

\subsection{A Practical Evaluation Scenario}
\label{sec: task incremental learning}
During training, the backbone network $\mathcal{M}$ sequentially learns the set of tasks $\mathcal{T}^t$, i.e., $T^{t_1}_1 \rightarrow T^{t_2}_2 \rightarrow 
... \rightarrow T^{t_M}_M$, in a task-incremental manner following the existing CL-based universal user representation learning approaches~\cite{kim2023task,yuan2021one}. 
However, the major difference between our proposed scenario and existing works~\cite{kim2023task,yuan2021one} is that \textit{while the same set of user behavior sequences is shared among all tasks in existing works, our work considers the expansion of user behavior sequences as time passes}. 
That is, for the same user, the length of the user behavior sequence may increase as time passes. That is, for user $u_l$ and its sequence $\mathbf{x}^{u_l}_{t_j}$ at time $t_j$, we have $|\mathbf{x}^{u_l}_{t_j}|\leq|\mathbf{x}^{u_l}_{t_{j+1}}|$. At the same time, the target labels $y^{T^{t_j}_i}_{u_l}$ for task $T_i$ are also different according to time $t_j$, i.e., for the same user $u_l$ and the same task $T_i$, we have $y^{T^{t_j}_i}_{u_l}\neq y^{T^{t_{j+1}}_i}_{u_l}$. For example, for the next item prediction task, the next item is different at time $t_j$ and $t_{j+1}$.

\smallskip
\noindent \textbf{Training of $T^{t_1}_1$:} In the first task, we learn the base user representation to be used in subsequent tasks. For this purpose, we train the user behavior sequence $\mathbf{x}^{u_l}_{t_1}$ autoregressively in a self-supervised manner to predict the next item as follows:
\begin{equation}
\footnotesize
    p(\mathbf{x}^{u_l}_{t_1};\Theta)=\prod_{r=1}^{|\mathbf{x}^{u_l}_{t_1}|-1} p(x_{r+1}^{u_l}|x_1^{u_l},...,x_{r}^{u_l}; \Theta),
    \label{Eq 3}
\end{equation}
where $p(x_{r+1}^{u_l}|x_1^{u_l},...,x_{r}^{u_l})$ denotes the probability of the $(r+1)$-th interaction of user $u_l$ given the user's prior interaction $\{x_1^{u_l},...,x_{r}^{u_l}\}$, and $\Theta$ comprises of the parameters of TCN. By training the TCN network to maximize the joint probability distribution of user $u_l$ as described above, we can obtain the base user representation that can be transferred to subsequent tasks.

\smallskip
\noindent \textbf{Training of $T^{t_j}_i$ ($i>1, j>1$):} After training the task $T^{t_1}_1$, we train subsequent tasks, e.g., $T^{t_1}_1 \rightarrow T^{t_2}_2 \rightarrow 
... \rightarrow T^{t_M}_M$, by utilizing the backbone network trained on task $T^{t_1}_1$. For the user behavior sequence $\mathbf{x}^{u_l}_{t_j}=\left\{x_1^{u_l},x_2^{u_l},...,x_r^{u_l}\right\}$, the backbone network produces autoregressive outputs, and thus the $r$-th row of the output $\mathbf{E}^{u_l}_K$, i.e., $\mathbf{E}^{u_l}_K[-1,:]\in\mathbb{R}^f$, represents the output of the network for a given behavior sequence $\left\{x_1^{u_l},x_2^{u_l},...,x_{r}^{u_l}\right\}$. $\mathbf{E}^{u_l}_K[-1,:]$ is considered as the representation of user $u_l$, and is used as input to the task-specific projector/classifier of task $T_i$, i.e., $G^{T_i}$, to predict the label (e.g., predicting the next favorite item, next purchase item, gender, etc) for user $u_l$ as follows:
\begin{equation}
\footnotesize
    \mathbf{\hat{y}}^{T^{t_j}_i}_{u_l} = \mathbf{E}_K^{u_l}[-1,:]\mathbf{W}^{T_i} + \mathbf{b}^{T_i} = G^{T_i}(\mathbf{E}_K^{u_l}), \quad {u_l \in \mathcal{U}^{T^{t_j}_i}}
\end{equation}
where $\mathbf{\hat{y}}^{T^{t_j}_i}_{u_l}$ denotes the predicted representation of $u_l$ for next item prediction task $T^{t_j}_i\in\mathcal{T}_{item}$, or the predicted label of $u_l$ for user profile prediction task $T^{t_j}_i\in\mathcal{T}_{profile}$. $\mathbf{W}^{T_i}$ and $\mathbf{b}^{T_i}$ represent the task-specific projection matrix and bias for task $T_i$, respectively.

\begin{figure}[t]
  \centering
  \includegraphics[width=1.00\linewidth]{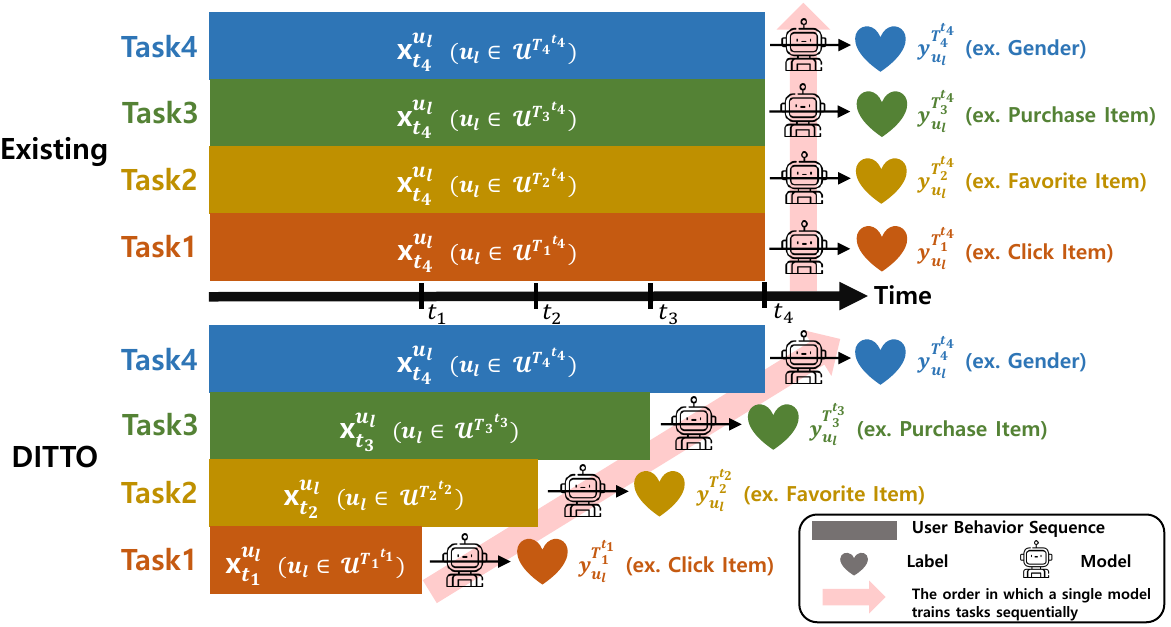}
  \caption{Train scenario of Existing Works and \proposed.}
  \label{fig:setting}
\end{figure}

\subsubsection{\textbf{Detailed Explanation for Experimental Scenario}}
Considering our assumption that the passage of time should be considered as tasks progress, we set different timestamps for each task during training. Figure~\ref{fig:setting} illustrates a toy example of the training scenario of the existing works~\cite{kim2023task,yuan2021one} and \proposed. Unlike the existing works, which perform training for all tasks at the same timestamp ($t_4$), indicating an unrealistic scenario where the time passage is not considered between tasks,
\proposed~trains $T^{t_1}_1$, i.e., predicting the next click item, using the user click sequence up to timestamp $t_1$ (i.e., $\mathbf{x}^{u_l}_{t_1}$, where $u_l \in \mathcal{U}^{T^{t_1}_1}$), and then trains $T^{t_2}_2$, i.e., predicting the next favorite item, using the extended user click sequence up to timestamp $t_2$ (i.e., $\mathbf{x}^{u_l}_{t_2}$, where $u_l \in \mathcal{U}^{T^{t_2}_2}$). Subsequently,~\proposed~gradually progresses the training with increasing timestamps for subsequent tasks. Finally, after being trained on the final task at $t_4$,~\proposed~performs inference on all the previous tasks.

\begin{table}[t]
\caption{The cosine similarity between the embeddings of the actual labels and the pseudo-labels generated with and without the distribution-aware sampling strategy. (Sec~\ref{sec:FKT_sampling})}
\resizebox{0.7\linewidth}{!}{
\begin{tabular}{c||ccccc}
\toprule
             &\multicolumn{5}{c}{Tmall}\\ \cline{2-6}
             & $T_1$ & $T_2$ & $T_3$ & $T_4$ & $T_5$  \\ \midrule\midrule
 \textit{w/o. sampling}    & -0.013 & 0.0271 & 0.0258 & 0.0277 & 0.0138 \\ \midrule
\textit{w. sampling}     & 0.3912 & 0.3413 & 0.3974 & 0.4078 & 0.3827 \\ \bottomrule
\end{tabular}}
\label{tab: pseudo_label}
\end{table}

\section{Proposed Method:~\proposed}

In this section, we describe our proposed method, \proposed, in detail. {Each module is developed considering the following two aspects: (1) Which user behavior sequences should be utilized for knowledge transfer? (2) How is the knowledge transferred?}

\subsection{Forward Knowledge Transfer: Alleviating Catastrophic Forgetting}

\label{sec:forward_knowledge_transfer} 

\subsubsection{Naive Pseudo-labeling}
To overcome catastrophic forgetting, the previous model's knowledge should be transferred to the current model (i.e., forward knowledge transfer). To this end, we duplicate the model into two parts: one is trainable to adapt to new tasks, and the other one is frozen and contains the knowledge learned from previous tasks. The trainable model is denoted by $\mathcal{M}(\cdot;\mathbf{m})$ and the frozen model is denoted as $\tilde{\mathcal{M}}(\cdot;\tilde{\mathbf{m}})$, and  $\mathbf{m}^{T_k}$ is the set of task-specific masks of a previous task $T_k$. Transferring the knowledge from previous tasks to the current task $T_i^{t_j}$ can be achieved through minimizing the following loss:

\begin{equation}
\footnotesize
    \mathcal{L}_{FKT} = \underset{u_l\in\mathcal{U}_{\textup{rand}}^{{T_i}^{t_j}}}{\mathbb{E}}\left[\underset{1\leq k < i}{\mathbb{E}}\left[{L_{{MSE}}}(\mathcal{M}(\mathbf{x}_{t_j}^{u_l};\mathbf{m}^{T_k}),\mathbf(\mathcal{\tilde{M}}(\mathbf{x}_{t_j}^{u_l};\mathbf{\tilde{m}}^{T_k})\right]\right]
\label{eq:knowledge_retention}
\end{equation}
where $L_\textit{MSE}$ is the Mean Squared Error loss, and $\mathcal{U}_{\textup{rand}}^{{T_i}^{t_j}}$ denotes a subset randomly sampled from the user set of current task $T_i^{t_j}$, i.e., $\mathcal{U}^{{T_i}^{t_j}}$. Note that $\mathcal{\tilde{M}}(\mathbf{x}^{u_l};\mathbf{\tilde{m}}^{T_k})$ is considered as pseudo-labels of $u_l\in\mathcal{U}_{\textup{rand}}^{{T_i}^{t_j}}$  in terms of a previous task $T_k$. 

However, under the practical scenario we propose, for users interacting with newly emerged items, the knowledge from previous tasks may be negatively transferred to the current task by optimizing Eq.~\ref{eq:knowledge_retention}, since these new items have not been seen when training on previous tasks. Hence, the model from previous tasks may not provide reliable output for these items, and potentially aggravating catastrophic forgetting. To validate the reliability of pseudo-labels, we calculate the cosine similarity between the embeddings of the actual labels and the pseudo-labels (Table~\ref{tab: pseudo_label}). We observe relatively lower cosine similarity than that of the pseudo-labels generated using the proposed sampling strategy discussed in Section~\ref{sec:FKT_sampling}. As a result, we argue that naively adapting a pseudo-labeling approach may result in negative transfer, which aggravates catastrophic forgetting.

\subsubsection{Distribution-aware User Sampling}
\label{sec:FKT_sampling}

Instead of randomly sampling users, we propose to sample users whose behavior sequence distribution is similar to the distribution of item embeddings learned during previous tasks.
{As depicted in Figure~\ref{fig:architecture} (a)}, the key idea is, while training the current task (e.g., $T^{t_j}_i$), to sample user behavior sequences to be utilized for forward knowledge transfer based on the similarity between the mean of item embeddings trained from a previous task (e.g., $T^{t_m}_{k}$ for $m<j$ and $k<i$) and the pseudo-label of each user behavior sequence calculated using the model learned from the previous task as follows:

\begin{equation}
\footnotesize
    \mathcal{U}^{T_k\rightarrow T_i}_{FKT} = \underset{u_l}{\textup{argmax}}^{(S_{i,k})}\textup{cos}\Big(\mathbf{\tilde{h}}^{T^{t_j}_k}_{u_l},\mathbb{E}[\mathbf{I}^{T_k}]\Big), \quad {u_l \in \mathcal{U}^{T^{t_j}_i}}
\label{eq:FWT_sampling}
\end{equation}

\noindent where $\textup{cos}(\cdot , \cdot)$ and $\mathbb{E}(\cdot)$ denote cosine similarity and expectation, respectively, $\mathbf{\tilde{h}}^{T^{t_j}_k}_{u_l}=\mathcal{\tilde{M}}(\mathbf{x}^{u_l}_{t_j};\mathbf{\tilde{m}}^{T_k})$ represents pseudo-label of $u_l\in\mathcal{U}^{T^{t_j}_i}$ in terms of a previous task $T_k$ utilizing the frozen backbone network (i.e., $\mathcal{\tilde{M}}$), and the task-specific mask (i.e., $\tilde{\mathbf{m}}^{T_k}$). $S_{i,k}$ is the number of user behavior sequences to be sampled, and $\mathbf{I}^{T_k}$ denotes the item embedding matrix trained in task $T_k$.
Note that $\mathbb{E}[\mathbf{I}^{T_k}]\in\mathbb{R}^f$ is the mean of the embedding vectors of items in task $T_k$, and we use it to represent the item distribution of task $T_k$ which can be directly compared with user representations. Here, the ratio of user behavior sequences to be sampled (i.e., $\rho_{i,k}$) is determined by taking into consideration the similarity between the task-specific masks as follows:

\begin{equation}
\footnotesize
    \rho_{i,k} = 1-\frac{1}{K}\sum^K_{d=1}\sigma(c\times\text{cos}(\mathbf{m}_{d}^{T_i},{\mathbf{\tilde{m}}}_{d}^{T_k})),
    \label{Eq:sampling}
\end{equation}
where $\rho_{i,k}$ represents the sampling rate of $T_i$ with respect to $T_k$ that specifies the amount of samples, i.e., $S_{i,k} = \rho_{i,k}\times|\mathcal{U}^{T^{t_j}_i}|$, $c$ is a scaling hyper-parameter, and ${\mathbf{\tilde{m}}}_{d}^{T_k}$ is the frozen mask of task $T_k$ before training the current task $T_i$. Note that $\rho_{i,k}$ is small if a previous task $T_k$ is similar to the current task $T_i$. 

\subsubsection{Distribution-aware Forward Knowledge Transfer}
Then, the pseudo-labels for the sampled user behavior sequences are obtained as follows:
\begin{equation}
\footnotesize
\mathbf{\tilde{h}}^{T^{t_j}_k}_{u_l} = \mathcal{\tilde{M}}(\mathbf{x}_{t_j}^{u_l};\mathbf{\tilde{m}}^{T_k}), \quad u_l\in\mathcal{U}^{T_k\rightarrow T_i}_{FKT},
\label{eq:pseudo_representation}
\end{equation}
where $\mathcal{\tilde{M}}$ and $\mathbf{\tilde{m}}^{T_k}$ indicate frozen backbone network and task-specific mask, respectively, which are used to generate the pseudo-labels that the current model needs to learn. It is important to note that the \textit{pseudo-labels can now be deemed reliable as they correspond to user behavior sequences exhibiting a distribution similar to the items observed during the training of a previous task $T_k$}, which is corroborated by result of \textit{w. sampling} in Table~\ref{tab: pseudo_label}. Given the pseudo-labels, we utilize the following loss to ensure that the current model learns the pseudo-labels, alleviating the forgetting of knowledge acquired from a previous task $T_k$ as follows:
\begin{equation}
\footnotesize
    \mathcal{L}^{T_k\rightarrow T_i}_{FKT} = \underset{u_l}{\mathbb{E}}\left[{L_{{MSE}}}(\mathcal{M}(\mathbf{x}_{t_j}^{u_l};\mathbf{m}^{T_k}),\mathbf{\tilde{h}}^{T^{t_j}_k}_{u_l})\right], \quad u_l\in\mathcal{U}^{T_k\rightarrow T_i}_{FKT}
\label{eq:FKT_loss}
\end{equation}
where $L_{{MSE}}$ is the Mean Squared Error loss. $\mathcal{M}$ and $\mathbf{m}^{T_k}$ are trained by optimizing the above loss, which allows the current model to retain the knowledge acquired from a previous task $T_k$. Furthermore, to ensure that the model retains the knowledge learned from all previous tasks (i.e., $T_{1:i}$\footnote{$T_{a:b}$ refers to the range from task $T_a$ to task $T_{b-1}$.}), we construct the final loss as follows:
\begin{equation}
\footnotesize
    \mathcal{L}_{FKT} = \underset{1\leq k < i}{\mathbb{E}}\left[\mathcal{L}^{T_k\rightarrow T_i}_{FKT}\right].
\label{eq:FKT_loss}
\end{equation}

\begin{figure}[t]
  \centering
  \includegraphics[width=1.0\linewidth]{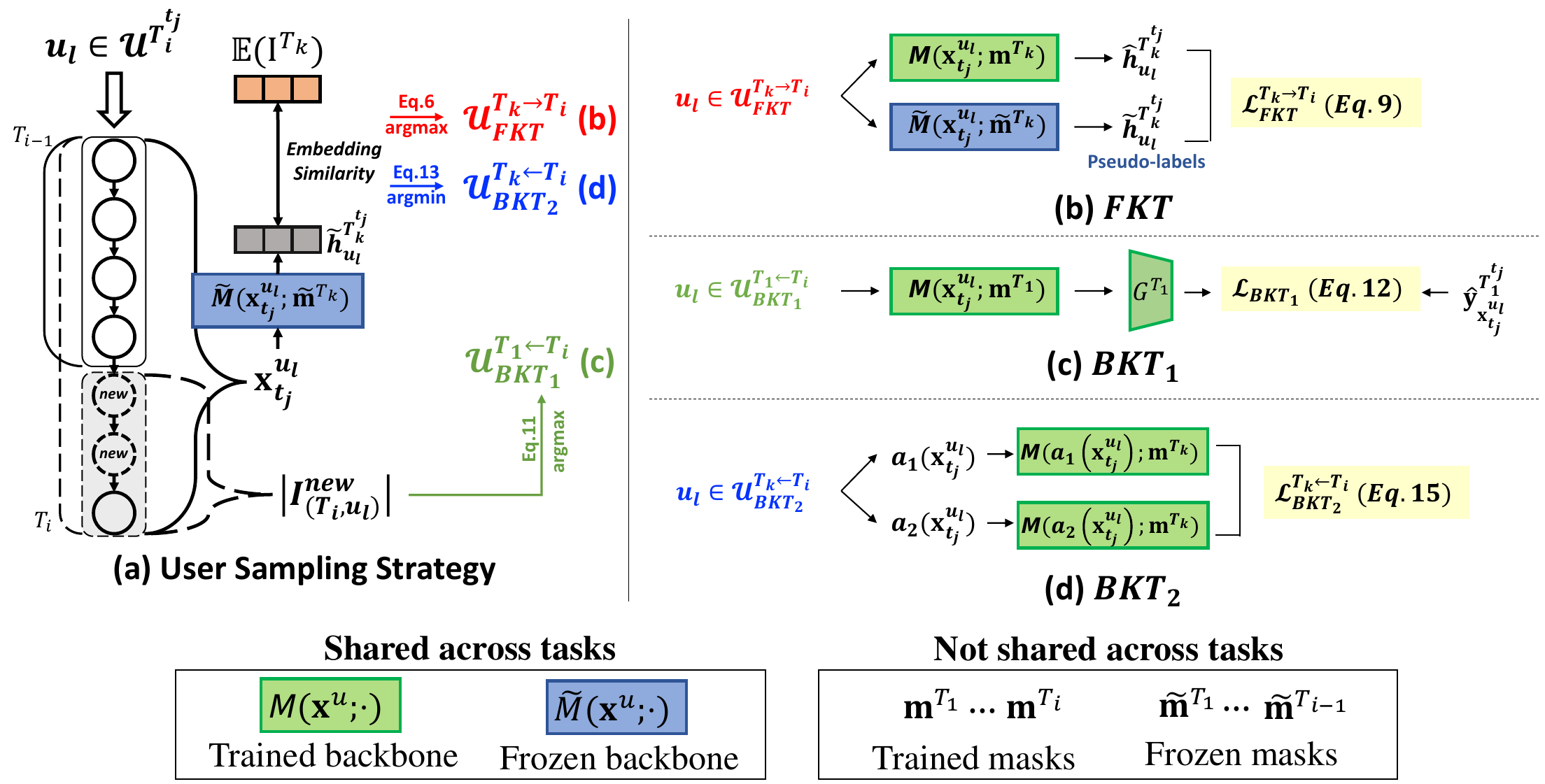}
  \caption{Overall model framework. (a) User sampling strategy for FKT, BKT$_1$, and BKT$_2$ modules in task $T_i^{t_j}$. (b) FKT from a previous task $T_{k}$. (c) BKT$_1$ to a previous task $T_1$. (d) BKT$_2$ to a previous task $T_{k}$.}
  \label{fig:architecture}
\end{figure}

{The mechanism of FKT from $T_k$ to $T_i$ is illustrated in Figure~\ref{fig:architecture} (b).}

\subsection{Backward Knowledge Transfer: Adapting to Current Shifted Item Distribution}
\label{sec:backward_knowledge_transfer}

Despite the forward knowledge transfer module effectively alleviating catastrophic forgetting (Sec.~\ref{sec:forward_knowledge_transfer}), it is crucial to design a model that allows previously acquired knowledge to adapt to the current shifted item distribution.
To do so, the current task (e.g., $T^{t_j}_i$) should transfer knowledge backward to the previous tasks (e.g., $T^{t_m}_{k}$ for $m<j$ and $k<i$). We elaborate on this for task $T_1$ (i.e., $T_1\leftarrow T_i$) in Sec.~\ref{sec:t1}, and task $T_{2:i}$ (i.e., $T_{2:i}\leftarrow T_i$) in Sec.~\ref{sec:t2} separately.

\subsubsection{Task $T_1$}
\label{sec:t1}
In task $T_1$, the user behavior sequences are trained autoregressively in a self-supervised manner, aiming to learn the base user representation. 

\noindent \textbf{Distribution-aware User Sampling.} We sample behavior sequences of users in the current task $T_i$ who has interacted with many newly emerged items as follows:
\begin{equation}
\footnotesize
    \mathcal{U}^{T_1\leftarrow T_i}_{BKT_1} = \underset{u_l}{\textup{argmax}}^{(S_{i,k})} |\mathcal{I}^{new}_{(T_i,u_l)}|, \quad {u_l \in \mathcal{U}^{T^{t_j}_i}},
\label{eq:BWT1_sampling}
\end{equation}
where $S_{i,k}$ is determined through Eq.~\ref{Eq:sampling}, and $\mathcal{I}^{new}_{(T_i,u_l)}$ denotes the set of items in the behavior sequence of user $u_l$ that have newly emerged during the transition from task $T_{i-1}$ to task $T_i$. 

\noindent \textbf{Learning of Base User Representation.} Then, the model is trained in an autoregressive manner as follows:
\begin{equation}
\scriptsize
    \mathcal{L}_{BKT_1} = \underset{u_l}{\mathbb{E}}\left[\underset{1\leq k \leq |\mathbf{x}^{u_l}_{t_j}|}{\mathbb{E}}\left[L_{\textup{BPR}}\Big(\hat{\textbf{y}}^{T^{t_j}_1}_{\textbf{x}^{u_l}_{\left[:k\right]}},\textbf{y}^{T^{t_j}_1}_{\textbf{x}^{u_l}_k}, \textbf{y}^{T^{t_j}_1,\textbf{-}}_{\textbf{x}^{u_l}_k}\Big)\right]\right], \quad u_l\in\mathcal{U}^{T_1\leftarrow T_i}_{BKT_1}
\label{eq:BKT1}
\end{equation}

where $L_{\textup{BPR}}(\mathbf{a},\mathbf{a}^+,\mathbf{a}^-)$ is the BPR loss~\cite{rendle2012bpr} defined as: $\log\sigma(s(\mathbf{a},\mathbf{a}^+)-s(\mathbf{a},\mathbf{a}^-))$ where $s(\cdot)$ is a scoring function, e.g., dot product.  
$\mathbf{\hat{y}}^{T^{t_j}_1}_{\mathbf{x}^{u_l}_{\left[:k\right]}}$\footnote{Previously, the predicted label for user $u_l$ was denoted as $\mathbf{\hat{y}}^{T^{t_j}_1}_{u_l}$, but we have chosen this expression to better illustrate the autoregressive training process.} denotes the predicted label for user $u_l$'s behavior sequence $\mathbf{x}^{u_l}_{\left[:k\right]}$ \Big(i.e., $\mathbf{\hat{y}}^{T^{t_j}_1}_{\mathbf{x}^{u_l}_{\left[:k\right]}}={G}^{T_1}(\mathcal{M}(\mathbf{x}^{u_l}_{\left[:k\right]};\mathbf{m}^{T_1}))$\Big), and $y^{T^{t_j}_1}_{\mathbf{x}^{u_l}_k}$, $y^{T^{t_j}_1,\mathbf{-}}_{\mathbf{x}^{u_l}_k}$ are the positive/negative items, respectively. By doing so, we can train the base user representation for the current item distribution.

\subsubsection{Task ${T}_{2:i}$}
\label{sec:t2}

Recall that the main goal of BKT is to allow the knowledge acquired from previous tasks to adapt to the current shifted item distribution.

\noindent \textbf{Distribution-aware User Sampling.} To this end,  
we first sample users in the current task $T_i$ whose behavior sequence distribution
has been shifted the most compared with that in the previous task $T_k$ as follows:

\begin{equation}
    \mathcal{U}^{T_k\leftarrow T_i}_{BKT_2} = \underset{u_l}{\textup{argmin}}^{(S_{i,k})}\textup{cos}\Big(\mathbf{\tilde{h}}^{T^{t_j}_k}_{u_l},\mathbb{E}[\mathbf{I}^{T_k}]\Big), \quad {u_l \in \mathcal{U}^{T^{t_j}_i}}.
\label{eq:BWT2_sampling}
\end{equation}
where the notations are similarly defined as in Eq.~\ref{eq:FWT_sampling}. 
Then, we update the backbone network (i.e., $\mathcal{M}$) as well as the task-specific mask of a previous task (i.e., $\mathbf{m}^{T_k}$) to make them aware of these users in $\mathcal{U}^{T_k\leftarrow T_i}_{BKT_2}$. In other words, we show the previously acquired knowledge how the shifted distribution looks like.

\noindent \textbf{Distribution-aware Backward Knowledge Transfer.} To do so, for each user ${u_l}$ in $\mathcal{U}^{T_k\leftarrow T_i}_{BKT_2}$, we augment its behavior sequence in various ways followed by contrastive learning to maximize the agreement between positive pairs, while minimizing the disagreement between negative pairs. 

\smallskip
\noindent\textit{1) Augmenting User Behavior Sequence. }
Given the user behavior sequence $\mathbf{x}_{t_j}^{u_l} (u_l \in \mathcal{U}^{T_k\leftarrow T_i}_{BKT_2})$, we adopt two widely used augmentation strategies (i.e., $a_1(\mathbf{x}_{t_j}^{u_l})$: Item Masking \&  $a_2(\mathbf{x}_{t_j}^{u_l})$: Item Substituting), as in \cite{xie2022contrastive}.

\smallskip
\noindent\textit{2) Contrastive Learning. }
Taking into account a mini-batch comprising $N$ users, we employ two augmentation operators on the behavior sequence of user $u_l$, resulting in $2N$ augmented sequences. For each user $u_l\in\mathcal{U}^{T_k\leftarrow T_i}_{BKT_2}$, we treat ($a_1(\mathbf{x}_{t_j}^{u_l})$, $a_2(\mathbf{x}_{t_j}^{u_l})$) as the positive pair, and the remaining $2(N-1)$ augmented examples as negative samples, i.e., $S^{-}$. Then the loss function $\mathcal{L}^{u_l}_{cl}$ for the user $u_l$ can be defined as follows:
\begin{equation}
\footnotesize
    \mathcal{L}^{u_l}_{cl} = -\textup{log}\frac{\textup{exp}\Big(\textup{sim}\big(a_1(\mathbf{x}_{t_j}^{u_l}), a_2(\mathbf{x}_{t_j}^{u_l}\big)\Big)}{\textup{exp}\Big(\textup{sim}\big(a_1(\mathbf{x}_{t_j}^{u_l}), a_2(\mathbf{x}_{t_j}^{u_l}\big)\Big)+\underset{s^{-}\in S^{-}}{\Sigma}\textup{exp}\Big(\textup{sim}\big(a_1(\mathbf{x}_{t_j}^{u_l}), s^{-}\big)\Big)},
\label{eq:contrastive}
\end{equation}
where $\textup{sim}(\cdot, \cdot)$ denotes the cosine similarity between the predicted representation of the model (i.e., {$\textup{sim}(\textbf{x}_1, \textbf{x}_2)=\textup{cos}(\mathcal{M}(\mathbf{x}_1;\mathbf{m}^{T_k}),$ $\mathcal{M}(\mathbf{x}_2;\mathbf{m}^{T_k}))$}). Then, the loss for all sampled users in $\mathcal{U}^{T_k\leftarrow T_i}_{BKT_2}$ is defined as follows:
\begin{equation}
\footnotesize
    \mathcal{L}^{T_k\leftarrow T_i}_{BKT_2} = \underset{u_l}{\mathbb{E}}\left[\mathcal{L}^{u_l}_{cl}\right], \quad u_l\in \mathcal{U}^{T_k\leftarrow T_i}_{BKT_2}
\label{eq:contrastive}
\end{equation}
Furthermore, to ensure that the model transfers the knowledge to all previous tasks $T_{2:i}$, the final loss is computed as follows:

\begin{equation}
\footnotesize
    \mathcal{L}_{BKT_2} = \underset{2\leq k < i}{\mathbb{E}}\left[\mathcal{L}^{T_k\leftarrow T_i}_{BKT_2}\right]
\label{eq:BKT}
\end{equation}
In this manner, previous tasks effectively adapts to current shifted item distribution via backward knowledge transfer. Note that BKT is only conducted on the next item prediction tasks (i.e., $\mathcal{T}_{item}\subset T_{2:i}$) excluding the user profile prediction tasks (i.e., $\mathcal{T}_{profile}\subset T_{2:i}$) due to two reasons: (1) The user profile prediction tasks are relatively easier compared to the next item prediction task, so that by learning the current item embedding through backward knowledge transfer for task $T_1$ through Eq.~\ref{eq:BKT1}, the model can sufficiently adapt to the current shifted item distribution. (2) Due to a significantly small number of labels in the user profile prediction tasks (e.g., 2 and 6 classes in the gender and age prediction tasks, respectively) compared with the next item prediction tasks, there is a higher likelihood of sampling bias (i.e., negative samples that actually share the same label). {The mechanism of the BKT$_2$ module from $T_i$ to $T_k$ is illustrated in Figure~\ref{fig:architecture} (d).}

\begin{table}[t]
\centering
\caption{Data Statistics ($|\mathcal{U}^{T_i}|$: num. users in $T_i$, $|\mathcal{Y}^{T_i}|$: num. unique labels in $T_i$).}
\renewcommand{\arraystretch}{1.1}
    \resizebox{1.0\linewidth}{!}{
    \begin{tabular}{wc{1cm}|wc{0.8cm}wc{0.5cm}|wc{0.8cm}wc{0.5cm}|wc{0.8cm}wc{0.5cm}|wc{0.8cm}wc{0.5cm}|wc{0.8cm}wc{0.5cm}|wc{0.8cm}wc{0.5cm}}
    \toprule
     \multirow{2}{*}{Dataset} & \multicolumn{2}{c|}{Task 1 ($T^{t_1}_1$)} & \multicolumn{2}{c|}{Task 2 ($T^{t_2}_2$)} & \multicolumn{2}{c|}{Task 3 ($T^{t_3}_3$)} & \multicolumn{2}{c|}{Task 4 ($T^{t_4}_4$)} & \multicolumn{2}{c|}{Task 5 ($T^{t_5}_5$)} & \multicolumn{2}{c}{Task 6 ($T^{t_6}_6$)}   \\ \cline{2-13}
    & \multicolumn{1}{c}{$|\mathcal{U}^{T^{t_1}_1}|$} & \multicolumn{1}{wc{0.5cm}|}{$|\mathcal{Y}^{T^{t_1}_1}|$} & \multicolumn{1}{c}{$|\mathcal{U}^{T^{t_2}_2}|$} & \multicolumn{1}{wc{0.5cm}|}{$|\mathcal{Y}^{T^{t_2}_2}|$} & \multicolumn{1}{c}{$|\mathcal{U}^{T^{t_3}_3}|$} & \multicolumn{1}{wc{0.5cm}|}{$|\mathcal{Y}^{T^{t_3}_3}|$} & \multicolumn{1}{c}{$|\mathcal{U}^{T^{t_4}_4}|$} & \multicolumn{1}{wc{0.5cm}|}{$|\mathcal{Y}^{T^{t_4}_4}|$} & \multicolumn{1}{c}{$|\mathcal{U}^{T^{t_5}_5}|$} & \multicolumn{1}{wc{0.5cm}|}{$|\mathcal{Y}^{T^{t_5}_5}|$} & \multicolumn{1}{c}{$|\mathcal{U}^{T^{t_6}_6}|$} & \multicolumn{1}{wc{0.5cm}}{$|\mathcal{Y}^{T^{t_6}_6}|$} \\ \midrule \midrule
    \multirow{2}{*}{Tmall} & \multicolumn{2}{c|}{Click} & \multicolumn{2}{c|}{Cart} & \multicolumn{2}{c|}{Purchase} & \multicolumn{2}{c|}{Favorite} & \multicolumn{2}{c|}{Age} & \multicolumn{2}{c}{Gender} \\ 
     & \multicolumn{1}{wc{1cm}}{355K} & \multicolumn{1}{c|}{525K} & 1.29K   & \multicolumn{1}{c|}{526K} & 65K & \multicolumn{1}{wc{1cm}|}{591K} & 54K & \multicolumn{1}{c|}{648K} & 393K & \multicolumn{1}{c|}{6} & 402K &  \multicolumn{1}{c}{2}\\ \midrule
    \multirow{2}{*}{ML} & \multicolumn{2}{c|}{Click} & \multicolumn{2}{c|}{4-star} & \multicolumn{2}{c|}{5-star} & \multicolumn{2}{c|}{\multirow{2}{*}{-}} & \multicolumn{2}{c|}{\multirow{2}{*}{-}} & \multicolumn{2}{c}{\multirow{2}{*}{-}} \\ 
     & 53K & \multicolumn{1}{c|}{4K} & 1.2K & \multicolumn{1}{c|}{4.5K} & 6.1K &  \multicolumn{1}{c|}{6.8K} & \multicolumn{2}{c|}{} & \multicolumn{2}{c|}{}   & \multicolumn{2}{c}{} \\ \midrule
     \multirow{2}{*}{Taobao} & \multicolumn{2}{c|}{Page View} & \multicolumn{2}{c|}{Cart} & \multicolumn{2}{c|}{Favorite} & \multicolumn{2}{c|}{Buy} & \multicolumn{2}{c|}{\multirow{2}{*}{-}} & \multicolumn{2}{c}{\multirow{2}{*}{-}} \\ 
     & 434K & \multicolumn{1}{c|}{1.47M} & 311K & \multicolumn{1}{c|}{1.63M} & 295K &  \multicolumn{1}{c|}{1.83M} & 321K &  \multicolumn{1}{c|}{2.00M} & \multicolumn{2}{c|}{}   & \multicolumn{2}{c}{} \\ \bottomrule
    \end{tabular}}
\label{tab: data_statistics}
\end{table}

\begin{table*}[t]
\begin{minipage}{0.65\linewidth}{
\caption{{Results over various tasks on Tmall, ML, and Taobao datasets.}}
\resizebox{0.95\linewidth}{!}{
\begin{tabular}{c||cccccc||ccc||cccc}
\toprule
             &\multicolumn{6}{c||}{Tmall} & \multicolumn{3}{c||}{ML} &\multicolumn{4}{c}{Taobao}\\ \cline{2-14}
             & $T_1$ & $T_2$ & $T_3$ & $T_4$ & $T_5$ & $T_6$ & $T_1$ & $T_2$ & $T_3$ & $T_1$ & $T_2$ & $T_3$ & $T_4$ \\ \midrule\midrule
SinMo & 0.3002 & 0.1032 & 0.2234 & 0.2135 & 0.3327 & 0.7432 & 0.3603 & 0.0804 & 0.4531 & 0.3340 & 0.2087 & 0.2417 & 0.3526 \\ \midrule
FineAll    & 0.3022 & 0.1101 & 0.1568 & 0.1116 & 0.2831 & 0.7219 & 0.3603 & 0.0671 &  0.4015 & 0.3340 & 0.3081 & 0.3288 & 0.3442 \\
PeterRec     & 0.3022 & 0.1096 & 0.1588 & 0.1135 & 0.3165 & 0.7458 & 0.3603 & 0.0785 &  0.4565 & 0.3340 & 0.3393 & 0.3268 & 0.3691 \\ \midrule
MTL          & -      & 0.1327 & 0.2392 & 0.1367 & 0.2674 & 0.7070 & -      & 0.0685 &  0.4261 & - & 0.2523 & 0.2737 & 0.3120 \\ \midrule 
Piggyback    & 0.3022 & 0.1106 & 0.0944 & 0.0680 & 0.2638 & 0.6993 & 0.3603 & 0.0741 & 0.4108 & 0.3340 & 0.1848 & 0.1800 & 0.2868 \\ 
HAT          & 0.3597 & 0.1712 & 0.1726 & 0.1555 & \textbf{0.3557} & 0.7378 & 0.3517 & 0.0518 & 0.4592 & 0.2439 & 0.3001 & 0.3599 & 0.3488 \\ 
CONURE       & 0.3366 & 0.1417 & 0.2145 & 0.1927 & 0.3103 & 0.6994 & 0.3635 & 0.0625 & 0.4779 & 0.3241 & 0.3349 & 0.3569 & 0.4469 \\
TERACON       & 0.3698 & 0.2002 & 0.2405 & 0.2462 & 0.3170 & 0.7411 & 0.3755 & 0.0816 & 0.5094 & 0.2803 & 0.3272 & 0.4843 & 0.4858 \\ \midrule
\proposed    & \textbf{0.6102} & \textbf{0.2764} & \textbf{0.3058} & \textbf{0.3345} & 0.3209 & \textbf{0.7496} & \textbf{0.4168} & \textbf{0.0890} & \textbf{0.5107} & \textbf{0.4291} & \textbf{0.3407} & \textbf{0.4903} & \textbf{0.4960}\\ \bottomrule
\end{tabular}}
\label{tab: main}}\end{minipage}
\begin{minipage}{0.34\linewidth}{
\caption{{Model performance with degradation ratio (in bracket) compared to Table \ref{tab: main} (\%) after training on a noisy task $T_{noise}$.}}
\resizebox{0.99\linewidth}{!}{
\begin{tabular}{c||ccccccc}
\toprule
        & \multicolumn{7}{c}{Tmall} \\ \cline{2-8}
        & $T_1$ & $T_2$ & $T_3$ & $T_{noise}$ & $T_4$ & $T_5$ & $T_6$ \\ \midrule \midrule
\multirow{2}{*}{HAT} & 0.3495 & 0.1612 & 0.1553 & \multirow{2}{*}{-} & 0.1021 & 0.3145 & 0.7211 \\ 
        & \small{(-2.84 \%)} & \small{(-5.84 \%)} & \small{(-10.02 \%)} &  & \small{(-34.34 \%)} & \small{(-11.58 \%)} & \small{(-2.26 \%)} \\ 
\multirow{2}{*}{CONURE} & 0.3366 & 0.1417 & 0.2145 & \multirow{2}{*}{-} & 0.1526 & 0.3020 & 0.6901 \\ 
        & \small{(0.0 \%)} & \small{(0.0 \%)} & \small{(0.0 \%)} &  & \small{(-20.81 \%)} & \small{(-2.67 \%)} & \small{(-1.33 \%)} \\
\multirow{2}{*}{TERACON} & 0.3458 & 0.1875 & 0.1951 & \multirow{2}{*}{-} & 0.2261 & 0.3004 & 0.7343 \\ 
        & \small{(-6.49 \%)} & \small{(-6.34 \%)} & \small{(-18.88 \%)} &  & \small{(-8.16 \%)} & \small{(-5.24 \%)} & \small{(-0.92 \%)} \\\midrule
\multirow{2}{*}{\proposed}   & \textbf{0.5967} & \textbf{0.2697} & \textbf{0.2854} & \multirow{2}{*}{-} & \textbf{0.3125} & \textbf{0.3137} & \textbf{0.7455} \\
        & \small{(-2.21 \%)} & \small{(-2.42 \%)} & \small{(-6.67 \%)} &  & \small{(-6.58 \%)} & \small{(-2.24 \%)} & \small{(-0.55 \%)} \\ \bottomrule

\end{tabular}}
\label{tab: noisy task}}\end{minipage}
\end{table*}
\subsection{Training and Inference}

\subsubsection{Training}

For a given task $T^{t_j}_i$, \proposed~is trained by optimizing the following loss:
\begin{equation}
    \mathcal{L} = \mathcal{L}_{main} + \alpha\mathcal{L}_{FKT} + \beta\mathcal{L}_{BKT_1} + \gamma\mathcal{L}_{BKT_2},
\label{eq:l_main}
\end{equation}
where $\alpha$, $\beta$, and $\gamma$ are the hyper-parameters which control the weight of $\mathcal{L}_{FKT}$, $\mathcal{L}_{BKT_1}$, and $\mathcal{L}_{BKT_2}$, respectively. The loss for each downstream task, i.e., $\mathcal{L}_{main}$, is defined based on the type of each task (i.e., $\mathcal{T}_{item}$ or $\mathcal{T}_{profile}$) as follows:
\begin{equation}
\footnotesize
    \mathcal{L}_{main} = \begin{cases}\underset{u_l}{\mathbb{E}}\left[L_{\textup{BPR}}\Big({G}^{T_i}(\mathcal{M}(\mathbf{x}_{t_j}^{u_l};\mathbf{m}^{T_i})), \textbf{y}^{T^{t_j}_i}_{u_l}, \textbf{y}^{T^{t_j}_i,\textbf{-}}_{u_l}\Big)\right], \; \text{for} \,\,u_l\in\mathcal{U}^{T^{t_j}_i(\in \mathcal{T}_{item})}, \\
    \underset{u_l}{\mathbb{E}}\left[L_{\textup{CE}}\Big({G}^{T_i}(\mathcal{M}(\mathbf{x}_{t_j}^{u_l};\mathbf{m}^{T_i})), \textbf{y}^{T^{t_j}_i}_{u_l}\Big)\right], \; \text{for} \,\,u_l\in\mathcal{U}^{T^{t_j}_i(\in \mathcal{T}_{profile})},
    \end{cases}
\label{eq:l_main_next_item}
\end{equation}
where $L_{BPR}$ and $L_{CE}$ are the BPR loss and cross-entropy loss, respectively, $\textbf{y}^{T^{t_j}_i}_{u_l}$ denotes ground truth label vector of user $u_l$ in task $T^{t_j}_i$, and $\textbf{y}^{T^{t_j}_i,\textbf{-}}_{u_l}$ denotes the label of a negative item for user $u_l$ in task $T^{t_j}_i$. Please note that above optimization process sequentially proceeds for the $M$ tasks in $\mathcal{T}$, i.e., $T^{t_1}_1 \rightarrow T^{t_2}_2 \rightarrow ... \rightarrow T^{t_M}_M$.

\subsubsection{Inference}

After completing the sequential training of all tasks, i.e., $T^{t_1}_1 \rightarrow T^{t_2}_2 \rightarrow 
... \rightarrow T^{t_M}_M$, we conduct inference on all the task in $\mathcal{T}$. In particular, for a given task $T_i\in\mathcal{T}$, we acquire the task-specific mask $\mathbf{m}^{T_i}$ and task-specific projector $G^{T_i}$. For each task $T_i\in\mathcal{T}$, \proposed~conducts inference with the backbone model $\mathcal{M}$ that is obtained after being trained on the last task $T^{t_M}_M$. Specifically, for task $T_i$, we make the model predictions for each user $u_l\in\mathcal{U}^{T^{t_M}_i}$, whose behavior sequence is given by $\mathbf{x}^{u_l}_{t_M}$, using the backbone network $\mathcal{M}$ as follows:
\begin{equation}
\footnotesize
    \hat{\textbf{y}}^{T^{t_M}_i}_{u_l} = {G}^{T_i}(\mathcal{M}(\mathbf{x}_{t_M}^{u_l};\mathbf{m}^{T_i})), \quad u_l\in\mathcal{U}^{T^{t_M}_i},
\end{equation}
where $\hat{\textbf{y}}^{T^{t_M}_i}_{u_l}$ is the predicted labels of $u_l$ in task $T^{t_M}_i$. 

\looseness=-1
It is important to note that in the training phase, user behavior sequences up to timestamp $t_i$ are used to train task $T_i$, i.e., $T_{i}^{t_i}$, while in the inference phase user behavior sequences up to the last timestamp $t_M$ are used to conduct inference on task $T_i$, i.e., $T_i^{t_M}$. We emphasize again that this is indeed a practical evaluation scenario on which CL-based universal user representation learning approaches should be evaluated. Figure~\ref{fig:setting} illustrates how training is performed.

\section{Experiments}

\noindent \textbf{Datasets and Task.} As~\proposed~ is the first work in continual learning designed and evaluated under a practical scenario that takes into account the passage of time as tasks progress, it is required to use datasets that contain (1) unique user IDs, (2) timestamps, and (3) multiple tasks. We were able to find three public datasets that satisfy the above criteria (i.e., Tmall, Movie Lens, and Taobao). 
We create various tasks from each public dataset in a similar manner to prior studies \cite{kim2023task, yuan2021one}. 
The detailed statistics for each dataset are presented in Table~\ref{tab: data_statistics}, and task descriptions are provided as follows:
\begin{itemize}[leftmargin=0.5cm]
\item \textbf{Tmall}\footnote{https://tianchi.aliyun.com/dataset/42} consists of four item recommendation tasks and two user profiling tasks. $T_1$ contains userID and the users' recent 100 item click interactions. Based on the users' interaction history, $T_2$, $T_3$, and $T_4$ aim to predict add-to-cart interactions, purchase interactions, and add-to-favorite interactions, respectively. Moreover, in $T_5$ and $T_6$, models are trained to predict user's age and gender, respectively. 
\item \textbf{MovieLens}\footnote{https://grouplens.org/datasets/movielens/} consists of three tasks, all of which are related to item recommendation. In particular, $T_1$ contains user IDs along with their most recent 30 click interactions, omitting items rated with 4-stars and 5-stars by the user. Using these recent 30 click interactions, in $T_2$ and $T_3$, models are trained to predict the items rated with 4-stars and 5-stars by the user, respectively.
\item \textbf{Taobao}\footnote{https://tianchi.aliyun.com/dataset/649} consists of four item recommendation tasks. $T_1$ contains userID and the users' recent 50 item page-view interactions. Using these recent 50 page-view interactions, in $T_2$, $T_3$, and $T_4$, models are trained to predict add-to-cart interactions, add-to-favorite interactions, and buy interactions, respectively. 
\end{itemize}

\smallskip
\noindent \textbf{Baselines.} We compare~\proposed~ with the most recent state-of-the-art CL-based universal user representation learning approaches, i.e., CONURE \cite{yuan2021one} and TERACON \cite{kim2023task}. We also include methods that 1) train a task-specific model (i.e., SinMO), 2) transfer learning approaches (i.e., FineAll an PeterRec \cite{yuan2020parameter}), multi-task learning approach (i.e., MTL), and continual learning approaches (i.e., Piggyback \cite{mallya2018piggyback}, HAT \cite{serra2018overcoming}, CONURE \cite{yuan2021one}, and TERACON \cite{kim2023task}). The detailed descriptions of each method are as follows:

\begin{itemize}[leftmargin=0.5cm]
    \item \textbf{SinMo} trains a separate model from scratch for each task, resulting in a total of $M$ SinMo models for the $M$ number of tasks. There is no transfer of knowledge between tasks in SinMo.
    \item \textbf{FineAll} is pre-trained on $T_1^{t_1}$ and then fine-tunes the model independently for each task, resulting in $M$ FineAll models for the $M$ number of tasks. Note that fine-tuning involves updating all parameters in the backbone network and creating task specific classifier $G^{T_i}$ from the scratch.
    \item \textbf{PeterRec} \cite{yuan2020parameter} is pre-trained on $T_1^{t_1}$ and then fine-tunes the only a small number of parameters called task-specific patches for each task, unlike FineAll. There are $M$ PeterRec models, one for each task.
    \item \textbf{MTL} trains the model using multi-task learning. Due to variations in the number of users across tasks, MTL employs two, i.e., one for $T_1^{t_j}$ and the other for $T_i^{t_j} (i>1, j>1)$, following a similar approach as in previous works \cite{kim2023task, yuan2021one}.
    \item \textbf{Piggyback} \cite{mallya2018piggyback} is a continual learning methods, which learns a binary mask for each task after obtaining the model parameters by pre-training on $T_1^{t_1}$.
    \item \textbf{HAT} \cite{serra2018overcoming} learns a soft mask for the output of each layer of the model to generate task specific output to conduct continual learning. 
    \item \textbf{CONURE} \cite{yuan2021one} learns universal user representation based on the parameter isolation approach  by retaining only a portion of parameters crucial for each task.
    \item \textbf{TERACON} \cite{kim2023task} is our main baseline, which learns relation-aware task-specific masks to capture the relationships between tasks while preventing catastrophic forgetting through knowledge retention.
    
\end{itemize}

\smallskip
\noindent \textbf{Evaluation Protocol.}
For evaluation, we randomly split each dataset in $T^{t_j}_i$ into train/validation/test data of $80/5/15\%$ following prior studies \cite{yuan2021one,kim2023task}.
We use Mean Reciprocal Rank ($MRR@5$) to measure the model performance on task $T_i\in \mathcal{T}_{item}$, and the classification accuracy, i.e., $Acc = \frac{\# \text{Correct predictions}}{\# \text{Total number of users}}$, for task $T_i\in \mathcal{T}_{profile}$. Note that the performance of CL-based methods, including Piggyback, HAT, CONURE, TERACON, and \proposed, for each task is evaluated after sequentially training from $T^{t_1}_1$ to $T^{t_M}_M$.\footnote{It is important to note that directly measuring catastrophic forgetting by comparing the performance between the train and test phases for each task is not feasible due to the temporal gap, causing variations in the datasets used for performance measurement between the train and test phases, making direct performance comparison in terms of catastrophic forgetting impossible. Hence, we exclude direct measurement of catastrophic forgetting.} Furthermore, we measure the degree of knowledge transfer between tasks in terms of knowledge transfer (KT). More precisely, we use $KT^{T_i}=\frac{R^{(T_M,T_i)}-\bar{R}^{T_i}}{\bar{R}^{T_i}}\times 100$, where $R^{(T_k,T_i)}$ represents the evaluated performance on $T_i$ after training on $T_k$, where $k>i$, and $\bar{R}^{T_i}$ represents the test performance of SinMo on $T_i$. $KT^{T_i}>0$ implies that the performance of $T_i$ after sequential training from $T_1$ to $T_M$ is superior compared with the case when it is evaluated after training a single model on $T_i$ from scratch. 

\begin{table*}[t]
  \caption{{Ablation study on Tmall dataset for each component of~\proposed. Each cell is designed to represent the performance of "integrated (users interacting only with existing items / users interacting with newly emerged items at least once)".}}
  \label{tab:ablation}
  \centering
  \resizebox{0.95\linewidth}{!}{
  \begin{tabular}{c|cc|c|c|c|c|c|c}
    \toprule
     \multirow{2}{*}{Row} & \multicolumn{2}{c|}{Component} & \multirow{2}{*}{\LARGE$T_1$} & \multirow{2}{*}{\LARGE$T_2$} & \multirow{2}{*}{\LARGE$T_3$} & \multirow{2}{*}{\LARGE$T_4$} & \multirow{2}{*}{\LARGE$T_5$} & \multirow{2}{*}{\LARGE$T_6$} \\
     \cline{2-3}
     & FKT & BKT & & & & & & \\
     \midrule
    (1) & \ding{55} & \ding{55}  & 0.2572 (0.2744 / 0.2517)   & 0.1994 (0.2056 / 0.1975) & 0.2309 (0.2487 / 0.2220) & 0.2170 (0.2217 / 0.2162) & 0.2281 (0.2317 / 0.2175) & 0.7446 (0.7446 / - )  \\
    (2) & \ding{51} & \ding{55}  & 0.2925 (0.3903 / 0.2627)  &  0.1955 (0.2248 / 0.1825) & 0.2371 (0.2685 / 0.2162)  & 0.2401 (0.2652 / 0.2226) & \textbf{0.3240} (0.3484 / 0.3016) & 0.7392 (0.7392 / - ) \\
    (3) & \ding{55} & \ding{51}  & 0.5717 (0.5624 / 0.5763)  & 0.2514 (0.2423 / 0.2524)  & 0.2799 (0.2780 / 0.2807)  & 0.3069 (0.3013 / 0.3105) & 0.3061 (0.3059 / 0.3067) & 0.7445 (0.7445 / - ) \\
    (4)-1 & \ding{51} & \ding{51} (only $\mathcal{L}_{BKT_1}$)  & 0.6073 (0.6092 / 0.6059)  & 0.2506 (0.2538 / 0.2475) & 0.2942 (0.3063 / 0.2869) & 0.3010 (0.3165 / 0.2886) & 0.3184 (0.3163 / 0.3251) & 0.7450 (0.7450 / - ) \\
    (4)-2 & \ding{51} & \ding{51} (only $\mathcal{L}_{BKT_2}$)  & 0.2816 (0.3878 / 0.2500)  & 0.2039 (0.2095 / 0.1984) & 0.2455 (0.2516 / 0.2322) & 0.2266 (0.2371 / 0.2233) & 0.2836 (0.2885 / 0.2792) & 0.7478 (0.7478 / - ) \\
    (5) & \ding{51} ($random$) & \ding{51} ($random$)  & 0.5968 (0.5804 / 0.6066)  & 0.1221 (0.1242 / 0.1190) & 0.1418 (0.1491 / 0.1288) & 0.1579 (0.1652 / 0.1437) & 0.3182 (0.3212 / 0.3173) & 0.7477 (0.7477 / - ) \\
    (6) & \ding{51} & \ding{51} & \textbf{0.6102} (0.6044 / 0.6099) & \textbf{0.2764} (0.2801 / 0.2722) & \textbf{0.3058} (0.3162 / 0.3013) & \textbf{0.3345} (0.3375 / 0.3338) & 0.3209 (0.3195 / 0.3256) & \textbf{0.7496} (0.7496 / - ) \\
  \bottomrule
\end{tabular}
}
\end{table*}

\begin{table*}[t]
\begin{minipage}{0.65\linewidth}{
\caption{{Model performance on Tmall dataset with original/reversed task sequence.}}
\resizebox{0.95 \linewidth}{!}{
\begin{tabular}{c||cc|cc|cc|cc|cc|cc}
\toprule
             \multirow{2}{*}{(a) Original} & \multicolumn{2}{c|}{$T_1$} & \multicolumn{2}{c|}{$T_2$} & \multicolumn{2}{c|}{$T_3$} & \multicolumn{2}{c|}{$T_4$} & \multicolumn{2}{c|}{$T_5$} & \multicolumn{2}{c}{$T_6$} \\ \cline{2-13}
             & MRR@5 & \multicolumn{1}{c|}{KT} & MRR@5 & \multicolumn{1}{c|}{KT} & MRR@5 & \multicolumn{1}{c|}{KT} & ACC & \multicolumn{1}{c|}{KT} & ACC & \multicolumn{1}{c|}{KT} & ACC & KT \\ \midrule \midrule
SinMo          & 0.3002 & - & 0.1032 & - & 0.2234 & - & 0.2135 & - & 0.3327 & - & 0.7432 & - \\ 
HAT          & 0.3597 & 19.82\% & 0.1712 & 65.89\% & 0.1726 & -22.74\% & 0.1555 & -27.17\% & 0.3557 & 6.91\% & 0.7378 & -0.73\% \\ 
CONURE       & 0.3366 & 12.13\% & 0.1417 & 37.31\% & 0.2145 & -3.98\% & 0.1927 & -9.74\% & 0.3103 & -6.73\% & 0.6994 & -5.89\% \\
TERACON       & 0.3698 & 23.18\% & 0.2002 & 93.99\% & 0.2405 & 7.65\% & 0.2462 & 15.32\% & 0.3170 & -4.72\% & 0.7411 & -0.28\% \\ \midrule
\proposed         & \textbf{0.6102} & 103.26\% & \textbf{0.2764} & 167.83\% & \textbf{0.3058} & 36.88\% & \textbf{0.3345} & 56.67\% & \textbf{0.3209} & -3.55\% & \textbf{0.7496} & 0.86\%\\ \midrule \midrule
             \multirow{2}{*}{(b) Reversed} & \multicolumn{2}{c|}{$T_1$} & \multicolumn{2}{c|}{$T_6$} & \multicolumn{2}{c|}{$T_5$} & \multicolumn{2}{c|}{$T_4$} & \multicolumn{2}{c|}{$T_3$} & \multicolumn{2}{c}{$T_2$} \\ \cline{2-13}
             & MRR@5 & \multicolumn{1}{c|}{KT} & ACC & \multicolumn{1}{c|}{KT} & ACC & \multicolumn{1}{c|}{KT} & ACC & \multicolumn{1}{c|}{KT} & MRR@5 & \multicolumn{1}{c|}{KT} & MRR@5 & KT \\ \midrule \midrule
SinMo          & 0.3222 & - & 0.7133 & - & \textbf{0.3768} & - & 0.2913 & - & 0.3360 & - & 0.4062 & - \\ 
HAT          & 0.4309 & 33.74\% & 0.6695 & -6.14\% & 0.3522 & -6.53\% & 0.4465 & 53.27\% & 0.4625 & 37.65\% & 0.4695 & 15.58\% \\ 
CONURE       & 0.3366 & 4.47\% & 0.7188 & 0.77\% & 0.3678 & -2.39\% & 0.2460 & -15.55\% & 0.3136 & -6.67\% & 0.4174 & 2.76\% \\
TERACON       & 0.4874 & 51.27\% & 0.6768 & -5.12\% & 0.3680 & -2.34\% & 0.4340 & 48.99\% & 0.4507 & 34.14\% & 0.4504 & 10.88\% \\ \midrule
\proposed         & \textbf{0.5430} & 68.53\% & \textbf{0.7678} & 7.64\% & 0.3700 & -1.80\% & \textbf{0.4816} & 65.33\% & \textbf{0.4677} & 39.20\% & \textbf{0.4710} & 15.95\% \\ \bottomrule
\end{tabular}}
\label{tab: task_order}}\end{minipage}
\begin{minipage}{0.34\linewidth}{
\caption{Model performance over various backbone networks on Tmall.}
\resizebox{0.95\linewidth}{!}{
\begin{tabular}{c||cc||cc}
\toprule
             & \multicolumn{2}{c||}{TERACON} &\multicolumn{2}{c}{\proposed}\\ \cline{1-5}
             Backbone & TCN & SASRec & TCN & SASRec \\ \midrule\midrule
$T_1$ & 0.3698 & 0.1958 & 0.6102 & 0.5299 \\
$T_2$ & 0.2002 & 0.1620 & 0.2764 & 0.2547 \\
$T_3$ & 0.2405 & 0.2154 & 0.3058 & 0.2811 \\
$T_4$ & 0.2462 & 0.1752 & 0.3345 & 0.3048 \\
$T_5$ & 0.3170 & 0.2660 & 0.3209 & 0.2687 \\
$T_6$ & 0.7411 & 0.7392 & 0.7496 & 0.7401 \\ \bottomrule
\end{tabular}}
\label{tab: backbone agnostic}
}\end{minipage}
\end{table*}

\subsection{Overall Performance}

The experimental results on three datasets are summarized in Table~\ref{tab: main}. We make the following key observations: 
\textbf{(1)} Transferring knowledge without considering the passage of time as in baseline models can lead to negative transfer (also refer to Table~\ref{tab: task_order} (a)). Comparing SinMo with TL-based models, i.e., FineAll and PeterRec, SinMo exhibits superior performance. This implies that when transferring knowledge from a model trained in the past, the model fails to learn knowledge about the current shifted item distribution, resulting in a degradation of performance. 
\textbf{(2)} CL-based methods, i.e., HAT, CONURE, and TERACON, outperform TL-based methods. This is attributed to the fact that, unlike TL-based methods, CL-based methods induce knowledge transfer not only between pairs of tasks but across multiple tasks. However, their performance is still lower than SinMo, which learns from scratch due to the inability of previous tasks to adapt to the current shifted item distribution. This becomes more evident from the performance differences across tasks. After initially learning the first task $T_1$, the number of newly emerged items is not substantial, and thus the item distribution undergoes minimal changes up to $T_2$. Consequently, TL-based and CL-based methods outperform SinMo during this period. 
However, as tasks progress, more new items emerge, leading to distribution shifts. Except for the relatively easier tasks (i.e., user profile prediction tasks), the performance of TL-based and CL-based methods becomes inferior to SinMo due to this distribution shift. \textbf{(3)} \proposed~outperforms baselines in the scenario where time passes as tasks progress. \proposed~achieves this by alleviating catastrophic forgetting through utilizing data that maintains past distributions through forward knowledge transfer, while also assisting previous knowledge in adapting to the current shifted item distribution by utilizing data with the shifted distribution through backward knowledge transfer. {\textbf{(4)} \proposed~demonstrates robust performance across various tasks on multiple datasets. The key factor of \proposed~is learning a universal user representation by utilizing the FKT and BKT modules to transfer knowledge bidirectionally at the representation layer before passing through the task-specific layer. Therefore, \proposed~can be generally applied to any backbone representation generation methods beyond TCN~\cite{yuan2019simple} as mentioned in Sec.~\ref{sec:backbone_network}. Refer to Sec.~\ref{sec: backbone agnostic} for more detail.

\subsection{Ablation Study}
\label{sec:ablation}
\looseness=-1
To comprehensively evaluate the impact of the FKT and BKT modules in~\proposed, we conduct ablation studies in Table~\ref{tab:ablation}. We conduct experiments on seven different cases, and the performance for each task is reported. 
Note that Row (6) is \proposed. We have the following observations:
\textbf{(1)} Introducing the FKT module is helpful (Row (1) vs. (2)). It is particularly more beneficial for users who interacted with existing items than those who interacted with newly emerged items. 
\textbf{(2)} introducing the BKT module (Row (3)) is particularly beneficial for users who have interacted with newly emerged items, since it helps previously acquired knowledge to adapt to the current shifted item distribution. 
\textbf{(3)} 
Upon closer analysis of the BKT component, through the sub-component $\mathcal{L}_{BKT_1}$ of the BKT module (Row (4)-1), transferring knowledge for $T_1$ leads to the learning of a base user representation for the current shifted item distribution, resulting in a reduction in forgetting for subsequent tasks. Solely utilizing the sub-component $\mathcal{L}_{BKT_2}$ (Row (4)-2) of the BKT module leads to an overall improvement in performance for previous tasks (Row (2) vs. Row (4)-2), however, it has a limitation in that it does not learn the base user representation through knowledge transfer for $T_1$. 
\textbf{(4)} Employing both the FKT module and the complete BKT module (Row (6)) shows superior performance, allowing the model to adapt to the current shifted item distribution while alleviating catastrophic forgetting of the knowledge acquired from previous tasks. 
\textbf{(5)} To verify the effectiveness of the user sampling strategy employed in the FKT and BKT modules, we randomly sampled users for each module (Row (5)). This results in generating unreliable pseudo-labels leading to negative forward transfer, and effective backward transfer is limited as it does not leverage users with shifted item distributions.


\subsection{Further Analysis}

\subsubsection{Order Robustness.}
As shown in Table~\ref{tab: task_order} (b), we conduct experiments by reversing the order of tasks excluding task $T_1$, i.e., $T^{t_1}_1\rightarrow T^{t_2}_6\rightarrow T^{t_3}_5\rightarrow T^{t_4}_4\rightarrow T^{t_5}_3\rightarrow T^{t_6}_2$. \proposed~maintains its superiority even when the order of tasks is changed. This verifies the effectiveness of maximizing positive knowledge transfer between tasks regardless of the task order. TERACON, due to the random sampling of user behavior sequences for the knowledge retention module, fails to generate reliable pseudo-labels for users with shifted item distributions (See Table~\ref{tab: pseudo_label}), leading to negative transfer. Additionally, CONURE, due to its parameter isolation-based nature, not only fails to adapt the model to the current shifted item distribution but also experiences a significant reduction in the available number of parameters as tasks progress, resulting in a decrease in knowledge transfer (KT) in later tasks compared to other CL-based methods.

\subsubsection{Noise Robustness.}
In Table~\ref{tab: noisy task}, we insert a noisy task between $T_3$ and $T_4$, i.e., $T^{t_1}_1\rightarrow T^{t_2}_2\rightarrow T^{t_3}_3\rightarrow T^{t_4}_{noise}\rightarrow T^{t_5}_4\rightarrow T^{t_6}_5\rightarrow T^{t_7}_6$.
We construct the noisy task by randomly assigning labels to the given user behavior sequences without noise. 
We observe that \proposed~exhibits less performance degradation compared with other CL-based methods, indicating its robustness to noise. The reasons \proposed~minimizes the adverse effects of noise labels are summarized as follows: \textbf{(1)} For the pseudo-labeling generation in the FKT module, \proposed~solely relies on the given user behavior sequences that are clean, while the noisy label information is neglected. Thus, previous knowledge is retained without being adversely affected noisy labels.
\textbf{(2)} Also, the BKT module transfers knowledge to previous tasks in a self-supervised manner by utilizing only the user behavior sequences without relying on the noisy label information. This makes the model avoid the adverse affect from noisy labels, and thus \proposed~transfers high-quality knowledge to the previous task, enabling the previous knowledge to adapt to the current shifted item distribution. In summary, \proposed~is robust to a noisy task as it does not rely on the label information that might be noisy, but solely relies on user behavior sequences for knowledge transfer.

\subsubsection{\proposed~is Backbone network-agnostic.} 
\label{sec: backbone agnostic}
To demonstrate that \proposed~is backbone network-agnostic, in Table~\ref{tab: backbone agnostic}, we compare the performance of TERACON and \proposed~with two different backbone networks $\mathcal{M}$, i.e., TCN \cite{yuan2019simple} and SASRec \cite{kang2018self} on the Tmall dataset. We observe that \proposed~outperforms TERACON regardless of the backbone network. This demonstrates that \proposed~is effective in practical scenarios by optimizing bidirectional positive knowledge transfer, irrespective of the backbone network employed for representing user behavior sequences.

\begin{table}[h]
\caption{Model performance with training time on Tmall.}
\resizebox{0.9\linewidth}{!}{
\begin{tabular}{c||cccccc||c}
\toprule
            &\multicolumn{6}{c||}{Tmall} & training time\\ \cline{2-7}
             & $T_1$ & $T_2$ & $T_3$ & $T_4$ & $T_5$ & $T_6$ & (min/epoch)  \\ \midrule\midrule
TERACON       & 0.3698 & 0.2002 & 0.2405 & 0.2462 & 0.3170 & 0.7411 & 26.75 \\ \midrule
\proposed       & 0.6102 & 0.2764 & 0.3058 & 0.3345 & 0.3209 & 0.7496 & 35.17 \\ \midrule
\proposed~(chunk)    & 0.5773 & 0.2596 & 0.2797 & 0.2965 & 0.3158 & 0.7402 & 31.18 \\ \bottomrule
\end{tabular}}
\label{tab: scalability_chunk}
\end{table}
\subsubsection{Scalability Analysis} {In Table~\ref{tab: scalability_chunk}, we compare the average training time between TERACON and \proposed. While \proposed~significant outperforms TERACON, \proposed~is only about 1.35 times slower than TERACON in training time. In addition, we conducted an additional experiment, \proposed~(chunk), which is an alternative strategy to reduce the cost of autoregressive training in the BKT$_1$ module. Specifically, it is a chunk-based training approach, which trains user behavior sequences every five steps instead of autoregressively at each step. This alternative allows for reduction in training time without significant performance loss compared to \proposed, serving as one possible solution for reducing training time.}

\begin{table}[t]
\caption{Model performance and training time (i.e., min/epoch) (in bracket) of \proposed~with and without sampling on Tmall dataset.}
\resizebox{0.9\linewidth}{!}{
\begin{tabular}{c||cccccc}
\toprule
    & $T_1$     & $T_2$     & $T_3$     & $T_4$     & $T_5$     & $T_6$     \\ \midrule\midrule
\multirow{2}{*}{\textit{sampling}} & 0.6102 & 0.2764 & 0.3058 & 0.3345 & 0.3209 & 0.7496 \\ 
    & \small{(13.14)} & \small{(0.05)} & \small{(2.41)} & \small{(5.64)} & \small{(11.46)} & \small{(13.85)} \\ \midrule
\textit{w/o} & 0.6801 & 0.2045 & 0.2890 & 0.3246 & 0.3309 & 0.7397 \\ 
\textit{sampling} & \small{(13.14)} & \small{(0.10)} & \small{(53.60)} & \small{(120.33)} & \small{(214.90)} & \small{(334.37)} \\ \bottomrule
\end{tabular}}
\label{tab: scalability_sample}
\end{table}


\subsubsection{Sampling Strategy}

In Table~\ref{tab: scalability_sample}, we additionally conduct experiments of the effect of sampling strategy of \proposed~on Tmall dataset. \textit{sampling} refers to the method that employs the sampling strategy for each component, as explained in Sec.~\ref{sec:FKT_sampling} (i.e., Eq.~\ref{eq:FWT_sampling}) and Sec.~\ref{sec:backward_knowledge_transfer} (i.e., Eq.~\ref{eq:BWT1_sampling}, Eq.~\ref{eq:BWT2_sampling}). \textit{w/o sampling} refers to the method that utilizes all user behavior sequences that have interacted only with existing items for the FKT module, and all user behavior sequences that have interacted with newly emerged items at least once for the BKT module.
\proposed~with sampling outperforms or performs in par with \proposed~w/o sampling in all tasks except for $T_1$, while \proposed~with sampling demonstrating 5-20 times faster training speed. This observation indicates that \proposed's sampling strategy is highly efficient and effective. We summarize the reasons as follows: \textbf{(1)} As $T_1$ can be trained autoregressively using only the user behavior sequence, the performance would increase just by having more user sequences involved in training. 
\textbf{(2)} However, for the remaining tasks, i.e., $T_{2:i}$, the choice of which users to sample has a significant impact on performance. In the FKT module, reliable pseudo-labels can only be generated from user behavior sequences that are similar to item distribution of previous tasks. Therefore, if pseudo-labels are generated from all users, unreliable pseudo-labels are utilized, leading to a negative transfer. \textbf{(3)} Additionally, the purpose of the BKT module for $T_{2:i}$ is to adapt the previously acquired knowledge to the current shifted item distribution. However, using all user behavior sequences indiscriminately can hinder the model's learning of knowledge about the current distribution, inevitably resulting in a performance decline. For these reasons, \proposed's sampling strategy excels in both effectiveness and speed of training.

\section{Conclusion}

In this paper, we introduce a  practical scenario on which CL-based universal user representation learning approaches should be evaluated, which takes into account the passage of time as tasks progress. Then, we propose a novel continual user representation learning framework, named \proposed, designed to alleviate catastrophic forgetting despite continuous shifts in item distribution, while also allowing the knowledge acquired from previous tasks to adapt to the current shifted item distribution.
\proposed~demonstrates promising performance across various real-world datasets under the practical evaluation scenario, and exhibits robust performance even with different task orders and the presence of noisy tasks, which highlights the applicability of \proposed~in real-world scenarios.

\looseness=-1
{There are a few limitations and possible extensions of our work. First, the sampling strategies in the FKT and BKT modules are solely based on the embedding similarity, which may not effectively represent the overall distribution. A possible direction could be to sample user behavior sequences by considering representativeness and diversity to better represent the past and current item distributions in FKT and BKT, respectively. Second, effectively training the model for cold items are challenging to achieve solely via knowledge transfer between tasks due to the lack of information. Therefore, we can extend the work to incorporate text modalities, such as item descriptions, to compensate for the relatively insufficient information.}

\begin{acks}
This work was supported by the National Research Foundation of Korea(NRF) grant funded by the Korea government(MSIT) (RS-2024-00335098), Institute of Information \& communications Technology Planning \& Evaluation (IITP) grant funded by the Korea government(MSIT) (RS-2022-II220157), and National Research Foundation of Korea(NRF) funded by Ministry of Science and ICT (NRF-2022M3J6A1063021).
\end{acks}

\bibliographystyle{ACM-Reference-Format}
\balance
\bibliography{acmart}


\begin{thebibliography}{36}


\ifx \showCODEN    \undefined \def \showCODEN     #1{\unskip}     \fi
\ifx \showISBNx    \undefined \def \showISBNx     #1{\unskip}     \fi
\ifx \showISBNxiii \undefined \def \showISBNxiii  #1{\unskip}     \fi
\ifx \showISSN     \undefined \def \showISSN      #1{\unskip}     \fi
\ifx \showLCCN     \undefined \def \showLCCN      #1{\unskip}     \fi
\ifx \shownote     \undefined \def \shownote      #1{#1}          \fi
\ifx \showarticletitle \undefined \def \showarticletitle #1{#1}   \fi
\ifx \showURL      \undefined \def \showURL       {\relax}        \fi
\providecommand\bibfield[2]{#2}
\providecommand\bibinfo[2]{#2}
\providecommand\natexlab[1]{#1}
\providecommand\showeprint[2][]{arXiv:#2}

\bibitem[Ba et~al\mbox{.}(2016)]%
        {ba2016layer}
\bibfield{author}{\bibinfo{person}{Jimmy~Lei Ba}, \bibinfo{person}{Jamie~Ryan Kiros}, {and} \bibinfo{person}{Geoffrey~E Hinton}.} \bibinfo{year}{2016}\natexlab{}.
\newblock \showarticletitle{Layer normalization}.
\newblock \bibinfo{journal}{\emph{arXiv preprint arXiv:1607.06450}} (\bibinfo{year}{2016}).
\newblock


\bibitem[Crawshaw(2020)]%
        {crawshaw2020multi}
\bibfield{author}{\bibinfo{person}{Michael Crawshaw}.} \bibinfo{year}{2020}\natexlab{}.
\newblock \showarticletitle{Multi-task learning with deep neural networks: A survey}.
\newblock \bibinfo{journal}{\emph{arXiv preprint arXiv:2009.09796}} (\bibinfo{year}{2020}).
\newblock


\bibitem[Gu et~al\mbox{.}(2021)]%
        {gu2021exploiting}
\bibfield{author}{\bibinfo{person}{Jie Gu}, \bibinfo{person}{Feng Wang}, \bibinfo{person}{Qinghui Sun}, \bibinfo{person}{Zhiquan Ye}, \bibinfo{person}{Xiaoxiao Xu}, \bibinfo{person}{Jingmin Chen}, {and} \bibinfo{person}{Jun Zhang}.} \bibinfo{year}{2021}\natexlab{}.
\newblock \showarticletitle{Exploiting behavioral consistence for universal user representation}. In \bibinfo{booktitle}{\emph{Proceedings of the AAAI Conference on Artificial Intelligence}}, Vol.~\bibinfo{volume}{35}. \bibinfo{pages}{4063--4071}.
\newblock


\bibitem[Guo et~al\mbox{.}(2017)]%
        {guo2017deepfm}
\bibfield{author}{\bibinfo{person}{Huifeng Guo}, \bibinfo{person}{Ruiming Tang}, \bibinfo{person}{Yunming Ye}, \bibinfo{person}{Zhenguo Li}, {and} \bibinfo{person}{Xiuqiang He}.} \bibinfo{year}{2017}\natexlab{}.
\newblock \showarticletitle{DeepFM: a factorization-machine based neural network for CTR prediction}.
\newblock \bibinfo{journal}{\emph{arXiv preprint arXiv:1703.04247}} (\bibinfo{year}{2017}).
\newblock


\bibitem[He and Chua(2017)]%
        {10.1145/3077136.3080777}
\bibfield{author}{\bibinfo{person}{Xiangnan He} {and} \bibinfo{person}{Tat-Seng Chua}.} \bibinfo{year}{2017}\natexlab{}.
\newblock \showarticletitle{Neural Factorization Machines for Sparse Predictive Analytics} \emph{(\bibinfo{series}{SIGIR '17})}. \bibinfo{publisher}{Association for Computing Machinery}, \bibinfo{address}{New York, NY, USA}, \bibinfo{pages}{355–364}.
\newblock
\showISBNx{9781450350228}
\href{https://doi.org/10.1145/3077136.3080777}{doi:\nolinkurl{10.1145/3077136.3080777}}


\bibitem[Hidasi et~al\mbox{.}(2015)]%
        {hidasi2015session}
\bibfield{author}{\bibinfo{person}{Bal{\'a}zs Hidasi}, \bibinfo{person}{Alexandros Karatzoglou}, \bibinfo{person}{Linas Baltrunas}, {and} \bibinfo{person}{Domonkos Tikk}.} \bibinfo{year}{2015}\natexlab{}.
\newblock \showarticletitle{Session-based recommendations with recurrent neural networks}.
\newblock \bibinfo{journal}{\emph{arXiv preprint arXiv:1511.06939}} (\bibinfo{year}{2015}).
\newblock


\bibitem[Kang and McAuley(2018)]%
        {kang2018self}
\bibfield{author}{\bibinfo{person}{Wang-Cheng Kang} {and} \bibinfo{person}{Julian McAuley}.} \bibinfo{year}{2018}\natexlab{}.
\newblock \showarticletitle{Self-attentive sequential recommendation}. In \bibinfo{booktitle}{\emph{2018 IEEE international conference on data mining (ICDM)}}. IEEE, \bibinfo{pages}{197--206}.
\newblock


\bibitem[Kim et~al\mbox{.}(2023a)]%
        {kim2023melt}
\bibfield{author}{\bibinfo{person}{Kibum Kim}, \bibinfo{person}{Dongmin Hyun}, \bibinfo{person}{Sukwon Yun}, {and} \bibinfo{person}{Chanyoung Park}.} \bibinfo{year}{2023}\natexlab{a}.
\newblock \showarticletitle{MELT: Mutual Enhancement of Long-Tailed User and Item for Sequential Recommendation}.
\newblock \bibinfo{journal}{\emph{arXiv preprint arXiv:2304.08382}} (\bibinfo{year}{2023}).
\newblock


\bibitem[Kim et~al\mbox{.}(2023b)]%
        {kim2023task}
\bibfield{author}{\bibinfo{person}{Sein Kim}, \bibinfo{person}{Namkyeong Lee}, \bibinfo{person}{Donghyun Kim}, \bibinfo{person}{Minchul Yang}, {and} \bibinfo{person}{Chanyoung Park}.} \bibinfo{year}{2023}\natexlab{b}.
\newblock \showarticletitle{Task Relation-aware Continual User Representation Learning}.
\newblock \bibinfo{journal}{\emph{arXiv preprint arXiv:2306.01792}} (\bibinfo{year}{2023}).
\newblock


\bibitem[Kirkpatrick et~al\mbox{.}(2017)]%
        {kirkpatrick2017overcoming}
\bibfield{author}{\bibinfo{person}{James Kirkpatrick}, \bibinfo{person}{Razvan Pascanu}, \bibinfo{person}{Neil Rabinowitz}, \bibinfo{person}{Joel Veness}, \bibinfo{person}{Guillaume Desjardins}, \bibinfo{person}{Andrei~A Rusu}, \bibinfo{person}{Kieran Milan}, \bibinfo{person}{John Quan}, \bibinfo{person}{Tiago Ramalho}, \bibinfo{person}{Agnieszka Grabska-Barwinska}, {et~al\mbox{.}}} \bibinfo{year}{2017}\natexlab{}.
\newblock \showarticletitle{Overcoming catastrophic forgetting in neural networks}.
\newblock \bibinfo{journal}{\emph{Proceedings of the national academy of sciences}} \bibinfo{volume}{114}, \bibinfo{number}{13} (\bibinfo{year}{2017}), \bibinfo{pages}{3521--3526}.
\newblock


\bibitem[Koren et~al\mbox{.}(2009)]%
        {5197422}
\bibfield{author}{\bibinfo{person}{Yehuda Koren}, \bibinfo{person}{Robert Bell}, {and} \bibinfo{person}{Chris Volinsky}.} \bibinfo{year}{2009}\natexlab{}.
\newblock \showarticletitle{Matrix Factorization Techniques for Recommender Systems}.
\newblock \bibinfo{journal}{\emph{Computer}} \bibinfo{volume}{42}, \bibinfo{number}{8} (\bibinfo{year}{2009}), \bibinfo{pages}{30--37}.
\newblock
\href{https://doi.org/10.1109/MC.2009.263}{doi:\nolinkurl{10.1109/MC.2009.263}}


\bibitem[Kumari et~al\mbox{.}(2022)]%
        {kumari2022retrospective}
\bibfield{author}{\bibinfo{person}{Lilly Kumari}, \bibinfo{person}{Shengjie Wang}, \bibinfo{person}{Tianyi Zhou}, {and} \bibinfo{person}{Jeff~A Bilmes}.} \bibinfo{year}{2022}\natexlab{}.
\newblock \showarticletitle{Retrospective adversarial replay for continual learning}.
\newblock \bibinfo{journal}{\emph{Advances in Neural Information Processing Systems}}  \bibinfo{volume}{35} (\bibinfo{year}{2022}), \bibinfo{pages}{28530--28544}.
\newblock


\bibitem[Li et~al\mbox{.}(2015)]%
        {10.1145/2806416.2806527}
\bibfield{author}{\bibinfo{person}{Sheng Li}, \bibinfo{person}{Jaya Kawale}, {and} \bibinfo{person}{Yun Fu}.} \bibinfo{year}{2015}\natexlab{}.
\newblock \showarticletitle{Deep Collaborative Filtering via Marginalized Denoising Auto-Encoder} \emph{(\bibinfo{series}{CIKM '15})}. \bibinfo{publisher}{Association for Computing Machinery}, \bibinfo{address}{New York, NY, USA}, \bibinfo{pages}{811–820}.
\newblock
\showISBNx{9781450337946}
\href{https://doi.org/10.1145/2806416.2806527}{doi:\nolinkurl{10.1145/2806416.2806527}}


\bibitem[Li and Zhao(2021)]%
        {10.5555/3491440.3492135}
\bibfield{author}{\bibinfo{person}{Sheng Li} {and} \bibinfo{person}{Handong Zhao}.} \bibinfo{year}{2021}\natexlab{}.
\newblock \showarticletitle{A Survey on Representation Learning for User Modeling}. In \bibinfo{booktitle}{\emph{Proceedings of the Twenty-Ninth International Joint Conference on Artificial Intelligence}} (Yokohama, Yokohama, Japan) \emph{(\bibinfo{series}{IJCAI'20})}. Article \bibinfo{articleno}{695}, \bibinfo{numpages}{7}~pages.
\newblock
\showISBNx{9780999241165}


\bibitem[Ma et~al\mbox{.}(2018)]%
        {10.1145/3219819.3220007}
\bibfield{author}{\bibinfo{person}{Jiaqi Ma}, \bibinfo{person}{Zhe Zhao}, \bibinfo{person}{Xinyang Yi}, \bibinfo{person}{Jilin Chen}, \bibinfo{person}{Lichan Hong}, {and} \bibinfo{person}{Ed~H. Chi}.} \bibinfo{year}{2018}\natexlab{}.
\newblock \showarticletitle{Modeling Task Relationships in Multi-Task Learning with Multi-Gate Mixture-of-Experts}. In \bibinfo{booktitle}{\emph{Proceedings of the 24th ACM SIGKDD International Conference on Knowledge Discovery \& Data Mining}} (London, United Kingdom) \emph{(\bibinfo{series}{KDD '18})}. \bibinfo{publisher}{Association for Computing Machinery}, \bibinfo{address}{New York, NY, USA}, \bibinfo{pages}{1930–1939}.
\newblock
\showISBNx{9781450355520}
\href{https://doi.org/10.1145/3219819.3220007}{doi:\nolinkurl{10.1145/3219819.3220007}}


\bibitem[Mallya et~al\mbox{.}(2018)]%
        {mallya2018piggyback}
\bibfield{author}{\bibinfo{person}{Arun Mallya}, \bibinfo{person}{Dillon Davis}, {and} \bibinfo{person}{Svetlana Lazebnik}.} \bibinfo{year}{2018}\natexlab{}.
\newblock \showarticletitle{Piggyback: Adapting a single network to multiple tasks by learning to mask weights}. In \bibinfo{booktitle}{\emph{Proceedings of the European Conference on Computer Vision (ECCV)}}. \bibinfo{pages}{67--82}.
\newblock


\bibitem[Mallya and Lazebnik(2018)]%
        {mallya2018packnet}
\bibfield{author}{\bibinfo{person}{Arun Mallya} {and} \bibinfo{person}{Svetlana Lazebnik}.} \bibinfo{year}{2018}\natexlab{}.
\newblock \showarticletitle{Packnet: Adding multiple tasks to a single network by iterative pruning}. In \bibinfo{booktitle}{\emph{Proceedings of the IEEE conference on Computer Vision and Pattern Recognition}}. \bibinfo{pages}{7765--7773}.
\newblock


\bibitem[McCloskey and Cohen(1989)]%
        {mccloskey1989catastrophic}
\bibfield{author}{\bibinfo{person}{Michael McCloskey} {and} \bibinfo{person}{Neal~J Cohen}.} \bibinfo{year}{1989}\natexlab{}.
\newblock \showarticletitle{Catastrophic interference in connectionist networks: The sequential learning problem}.
\newblock In \bibinfo{booktitle}{\emph{Psychology of learning and motivation}}. Vol.~\bibinfo{volume}{24}. \bibinfo{publisher}{Elsevier}, \bibinfo{pages}{109--165}.
\newblock


\bibitem[Nair and Hinton(2010)]%
        {nair2010rectified}
\bibfield{author}{\bibinfo{person}{Vinod Nair} {and} \bibinfo{person}{Geoffrey~E Hinton}.} \bibinfo{year}{2010}\natexlab{}.
\newblock \showarticletitle{Rectified linear units improve restricted boltzmann machines}. In \bibinfo{booktitle}{\emph{Proceedings of the 27th international conference on machine learning (ICML-10)}}. \bibinfo{pages}{807--814}.
\newblock


\bibitem[Ni et~al\mbox{.}(2018a)]%
        {Ni2018PerceiveYU}
\bibfield{author}{\bibinfo{person}{Yabo Ni}, \bibinfo{person}{Dan Ou}, \bibinfo{person}{Shichen Liu}, \bibinfo{person}{Xiang Li}, \bibinfo{person}{Wenwu Ou}, \bibinfo{person}{Anxiang Zeng}, {and} \bibinfo{person}{Luo Si}.} \bibinfo{year}{2018}\natexlab{a}.
\newblock \showarticletitle{Perceive Your Users in Depth: Learning Universal User Representations from Multiple E-commerce Tasks}.
\newblock \bibinfo{journal}{\emph{Proceedings of the 24th ACM SIGKDD International Conference on Knowledge Discovery \& Data Mining}} (\bibinfo{year}{2018}).
\newblock


\bibitem[Ni et~al\mbox{.}(2018b)]%
        {ni2018perceive}
\bibfield{author}{\bibinfo{person}{Yabo Ni}, \bibinfo{person}{Dan Ou}, \bibinfo{person}{Shichen Liu}, \bibinfo{person}{Xiang Li}, \bibinfo{person}{Wenwu Ou}, \bibinfo{person}{Anxiang Zeng}, {and} \bibinfo{person}{Luo Si}.} \bibinfo{year}{2018}\natexlab{b}.
\newblock \showarticletitle{Perceive your users in depth: Learning universal user representations from multiple e-commerce tasks}. In \bibinfo{booktitle}{\emph{Proceedings of the 24th ACM SIGKDD International Conference on Knowledge Discovery \& Data Mining}}. \bibinfo{pages}{596--605}.
\newblock


\bibitem[Ratcliff(1990)]%
        {ratcliff1990connectionist}
\bibfield{author}{\bibinfo{person}{Roger Ratcliff}.} \bibinfo{year}{1990}\natexlab{}.
\newblock \showarticletitle{Connectionist models of recognition memory: constraints imposed by learning and forgetting functions.}
\newblock \bibinfo{journal}{\emph{Psychological review}} \bibinfo{volume}{97}, \bibinfo{number}{2} (\bibinfo{year}{1990}), \bibinfo{pages}{285}.
\newblock


\bibitem[Rendle et~al\mbox{.}(2012)]%
        {rendle2012bpr}
\bibfield{author}{\bibinfo{person}{Steffen Rendle}, \bibinfo{person}{Christoph Freudenthaler}, \bibinfo{person}{Zeno Gantner}, {and} \bibinfo{person}{Lars Schmidt-Thieme}.} \bibinfo{year}{2012}\natexlab{}.
\newblock \showarticletitle{BPR: Bayesian personalized ranking from implicit feedback}.
\newblock \bibinfo{journal}{\emph{arXiv preprint arXiv:1205.2618}} (\bibinfo{year}{2012}).
\newblock


\bibitem[Ruder(2017)]%
        {ruder2017overview}
\bibfield{author}{\bibinfo{person}{Sebastian Ruder}.} \bibinfo{year}{2017}\natexlab{}.
\newblock \showarticletitle{An overview of multi-task learning in deep neural networks}.
\newblock \bibinfo{journal}{\emph{arXiv preprint arXiv:1706.05098}} (\bibinfo{year}{2017}).
\newblock


\bibitem[Serra et~al\mbox{.}(2018)]%
        {serra2018overcoming}
\bibfield{author}{\bibinfo{person}{Joan Serra}, \bibinfo{person}{Didac Suris}, \bibinfo{person}{Marius Miron}, {and} \bibinfo{person}{Alexandros Karatzoglou}.} \bibinfo{year}{2018}\natexlab{}.
\newblock \showarticletitle{Overcoming catastrophic forgetting with hard attention to the task}. In \bibinfo{booktitle}{\emph{International Conference on Machine Learning}}. PMLR, \bibinfo{pages}{4548--4557}.
\newblock


\bibitem[Wang et~al\mbox{.}(2019)]%
        {wang2019neural}
\bibfield{author}{\bibinfo{person}{Xiang Wang}, \bibinfo{person}{Xiangnan He}, \bibinfo{person}{Meng Wang}, \bibinfo{person}{Fuli Feng}, {and} \bibinfo{person}{Tat-Seng Chua}.} \bibinfo{year}{2019}\natexlab{}.
\newblock \showarticletitle{Neural graph collaborative filtering}. In \bibinfo{booktitle}{\emph{Proceedings of the 42nd international ACM SIGIR conference on Research and development in Information Retrieval}}. \bibinfo{pages}{165--174}.
\newblock


\bibitem[Wu et~al\mbox{.}(2019)]%
        {wu2019session}
\bibfield{author}{\bibinfo{person}{Shu Wu}, \bibinfo{person}{Yuyuan Tang}, \bibinfo{person}{Yanqiao Zhu}, \bibinfo{person}{Liang Wang}, \bibinfo{person}{Xing Xie}, {and} \bibinfo{person}{Tieniu Tan}.} \bibinfo{year}{2019}\natexlab{}.
\newblock \showarticletitle{Session-based recommendation with graph neural networks}. In \bibinfo{booktitle}{\emph{Proceedings of the AAAI conference on artificial intelligence}}, Vol.~\bibinfo{volume}{33}. \bibinfo{pages}{346--353}.
\newblock


\bibitem[Xie et~al\mbox{.}(2022)]%
        {xie2022contrastive}
\bibfield{author}{\bibinfo{person}{Xu Xie}, \bibinfo{person}{Fei Sun}, \bibinfo{person}{Zhaoyang Liu}, \bibinfo{person}{Shiwen Wu}, \bibinfo{person}{Jinyang Gao}, \bibinfo{person}{Jiandong Zhang}, \bibinfo{person}{Bolin Ding}, {and} \bibinfo{person}{Bin Cui}.} \bibinfo{year}{2022}\natexlab{}.
\newblock \showarticletitle{Contrastive learning for sequential recommendation}. In \bibinfo{booktitle}{\emph{2022 IEEE 38th international conference on data engineering (ICDE)}}. IEEE, \bibinfo{pages}{1259--1273}.
\newblock


\bibitem[Yang et~al\mbox{.}(2021)]%
        {yang2021autoft}
\bibfield{author}{\bibinfo{person}{Xiangli Yang}, \bibinfo{person}{Qing Liu}, \bibinfo{person}{Rong Su}, \bibinfo{person}{Ruiming Tang}, \bibinfo{person}{Zhirong Liu}, {and} \bibinfo{person}{Xiuqiang He}.} \bibinfo{year}{2021}\natexlab{}.
\newblock \showarticletitle{Autoft: Automatic fine-tune for parameters transfer learning in click-through rate prediction}.
\newblock \bibinfo{journal}{\emph{arXiv preprint arXiv:2106.04873}} (\bibinfo{year}{2021}).
\newblock


\bibitem[Yoon et~al\mbox{.}(2017)]%
        {yoon2017lifelong}
\bibfield{author}{\bibinfo{person}{Jaehong Yoon}, \bibinfo{person}{Eunho Yang}, \bibinfo{person}{Jeongtae Lee}, {and} \bibinfo{person}{Sung~Ju Hwang}.} \bibinfo{year}{2017}\natexlab{}.
\newblock \showarticletitle{Lifelong learning with dynamically expandable networks}.
\newblock \bibinfo{journal}{\emph{arXiv preprint arXiv:1708.01547}} (\bibinfo{year}{2017}).
\newblock


\bibitem[Yuan et~al\mbox{.}(2020)]%
        {yuan2020parameter}
\bibfield{author}{\bibinfo{person}{Fajie Yuan}, \bibinfo{person}{Xiangnan He}, \bibinfo{person}{Alexandros Karatzoglou}, {and} \bibinfo{person}{Liguang Zhang}.} \bibinfo{year}{2020}\natexlab{}.
\newblock \showarticletitle{Parameter-efficient transfer from sequential behaviors for user modeling and recommendation}. In \bibinfo{booktitle}{\emph{Proceedings of the 43rd International ACM SIGIR conference on research and development in Information Retrieval}}. \bibinfo{pages}{1469--1478}.
\newblock


\bibitem[Yuan et~al\mbox{.}(2019)]%
        {yuan2019simple}
\bibfield{author}{\bibinfo{person}{Fajie Yuan}, \bibinfo{person}{Alexandros Karatzoglou}, \bibinfo{person}{Ioannis Arapakis}, \bibinfo{person}{Joemon~M Jose}, {and} \bibinfo{person}{Xiangnan He}.} \bibinfo{year}{2019}\natexlab{}.
\newblock \showarticletitle{A simple convolutional generative network for next item recommendation}. In \bibinfo{booktitle}{\emph{Proceedings of the twelfth ACM international conference on web search and data mining}}. \bibinfo{pages}{582--590}.
\newblock


\bibitem[Yuan et~al\mbox{.}(2021)]%
        {yuan2021one}
\bibfield{author}{\bibinfo{person}{Fajie Yuan}, \bibinfo{person}{Guoxiao Zhang}, \bibinfo{person}{Alexandros Karatzoglou}, \bibinfo{person}{Joemon Jose}, \bibinfo{person}{Beibei Kong}, {and} \bibinfo{person}{Yudong Li}.} \bibinfo{year}{2021}\natexlab{}.
\newblock \showarticletitle{One person, one model, one world: Learning continual user representation without forgetting}. In \bibinfo{booktitle}{\emph{Proceedings of the 44th International ACM SIGIR Conference on Research and Development in Information Retrieval}}. \bibinfo{pages}{696--705}.
\newblock


\bibitem[Zhao et~al\mbox{.}(2013)]%
        {10.5555/2891460.2891628}
\bibfield{author}{\bibinfo{person}{Lili Zhao}, \bibinfo{person}{Sinno~Jialin Pan}, \bibinfo{person}{Evan~Wei Xiang}, \bibinfo{person}{Erheng Zhong}, \bibinfo{person}{Zhongqi Lu}, {and} \bibinfo{person}{Qiang Yang}.} \bibinfo{year}{2013}\natexlab{}.
\newblock \showarticletitle{Active Transfer Learning for Cross-System Recommendation}. In \bibinfo{booktitle}{\emph{Proceedings of the Twenty-Seventh AAAI Conference on Artificial Intelligence}} (Bellevue, Washington) \emph{(\bibinfo{series}{AAAI'13})}. \bibinfo{publisher}{AAAI Press}, \bibinfo{pages}{1205–1211}.
\newblock


\bibitem[Zhou and Cao(2021)]%
        {zhou2021overcoming}
\bibfield{author}{\bibinfo{person}{Fan Zhou} {and} \bibinfo{person}{Chengtai Cao}.} \bibinfo{year}{2021}\natexlab{}.
\newblock \showarticletitle{Overcoming catastrophic forgetting in graph neural networks with experience replay}. In \bibinfo{booktitle}{\emph{Proceedings of the AAAI Conference on Artificial Intelligence}}, Vol.~\bibinfo{volume}{35}. \bibinfo{pages}{4714--4722}.
\newblock


\bibitem[Zhou et~al\mbox{.}(2019)]%
        {zhou2019deep}
\bibfield{author}{\bibinfo{person}{Guorui Zhou}, \bibinfo{person}{Na Mou}, \bibinfo{person}{Ying Fan}, \bibinfo{person}{Qi Pi}, \bibinfo{person}{Weijie Bian}, \bibinfo{person}{Chang Zhou}, \bibinfo{person}{Xiaoqiang Zhu}, {and} \bibinfo{person}{Kun Gai}.} \bibinfo{year}{2019}\natexlab{}.
\newblock \showarticletitle{Deep interest evolution network for click-through rate prediction}. In \bibinfo{booktitle}{\emph{Proceedings of the AAAI conference on artificial intelligence}}, Vol.~\bibinfo{volume}{33}. \bibinfo{pages}{5941--5948}.
\newblock


\end{thebibliography}

\end{document}